\documentclass[10pt,twocolumn,letterpaper]{article}

\usepackage{cvpr}
\usepackage{times}
\usepackage{epsfig}
\usepackage{graphicx}
\usepackage{amsmath}
\usepackage{amssymb}
\usepackage{tabularx}
\usepackage{booktabs}
\usepackage{enumerate}
\usepackage{array,multirow}

\usepackage[pagebackref=true,breaklinks=true,letterpaper=true,colorlinks,bookmarks=false]{hyperref}

\cvprfinalcopy %

\newcommand{\PAR}[1]{\vskip4pt \noindent {\bf #1~}}

\DeclareMathOperator{\track}{\mathbf{X}} %
\DeclareMathOperator{\bbox}{\mathbf{x}} %
\DeclareMathOperator{\obj}{\mathbf{y}}
\DeclareMathOperator{\distmat}{\mathbf{D}}
\DeclareMathOperator{\assigmat}{\mathbf{A}}

\DeclareMathOperator{\binmat}{\mathbf{B}}
\DeclareMathOperator{\conmat}{\mathbf{C}}
\DeclareMathOperator{\softassigmat}{\tilde{\mathbf{A}}}

\DeclareMathOperator{\tp}{\text{TP}}
\DeclareMathOperator{\fp}{\text{FP}}
\DeclareMathOperator{\fn}{\text{FN}}
\DeclareMathOperator{\ids}{\text{IDS}}

\DeclareMathOperator{\difffp}{\tilde{\text{FP}}}
\DeclareMathOperator{\difffn}{\tilde{\text{FN}}}
\DeclareMathOperator{\diffids}{\tilde{\text{IDS}}}

\newcommand{\R}{\mathbb{R}}

\newcommand{\diffmota}{dMOTA}
\newcommand{\diffmotp}{dMOTP}

\newcommand{\method}{DeepMOT}

\begin{document}

\title{How To Train Your Deep Multi-Object Tracker\vspace{-3mm}}

\author{Yihong Xu$^1$
\quad
Aljo\u sa O\u sep$^2$
\quad
Yutong Ban$^{1,3}$
\quad
Radu Horaud$^1$\\
\quad
Laura Leal-Taix\'{e}$^2$
\quad
Xavier Alameda-Pineda$^1$\\
$^1$Inria, LJK, Univ. Grenoble Alpes, France \quad $^2$Technical University of Munich, Germany\\ $^3$Distributed Robotics Lab, CSAIL, MIT, USA\\
$^1${\tt\small \{firstname.lastname\}@inria.fr} \quad
$^2${\tt\small \{aljosa.osep, leal.taixe\}@tum.de} \quad
$^3${\tt\small yban@csail.mit.edu}
}

\maketitle

\begin{abstract}
The recent trend in vision-based multi-object tracking (MOT) is heading towards leveraging the representational power of deep learning to jointly learn to detect and track objects. However, existing methods train only certain sub-modules using loss functions that often do not correlate with established tracking evaluation measures such as Multi-Object Tracking Accuracy (MOTA) and Precision (MOTP). As these measures are not differentiable, the choice of appropriate loss functions for end-to-end training of multi-object tracking methods is still an open research problem. In this paper, we bridge this gap by proposing a differentiable proxy of MOTA and MOTP, which we combine in a loss function suitable for end-to-end training of deep multi-object trackers. As a key ingredient, we propose a Deep Hungarian Net (DHN) module that approximates the Hungarian matching algorithm. DHN allows estimating the correspondence between object tracks and ground truth objects to compute differentiable proxies of MOTA and MOTP, which are in turn used to optimize deep trackers directly. We experimentally demonstrate that the proposed differentiable framework improves the performance of existing multi-object trackers, and we establish a new state of the art on the MOTChallenge benchmark. Our code is publicly available from \href{https://github.com/yihongXU/deepMOT}{https://github.com/yihongXU/deepMOT}.
\end{abstract}
\vspace{-6mm}

\section{Introduction}
\vspace{-2mm}
Vision-based multi-object tracking (MOT) is a long-standing research problem with applications in mobile robotics and autonomous driving. It is through tracking that we become aware of surrounding object instances and anticipate their future motion.
The majority of existing methods for pedestrian tracking follow the tracking-by-detection paradigm and mainly focus on the association of detector responses over time.
A significant amount of research investigated combinatorial optimization techniques for this challenging data association problem~\cite{Reid79TAC, Pirsiavash11CVPR, Schulter17CVPR, Zhang08CVPR, ButtCollins13CVPR, Brendel11CVPR}.

\begin{figure}[t]
    \centering
    \includegraphics[width=1.0\linewidth]{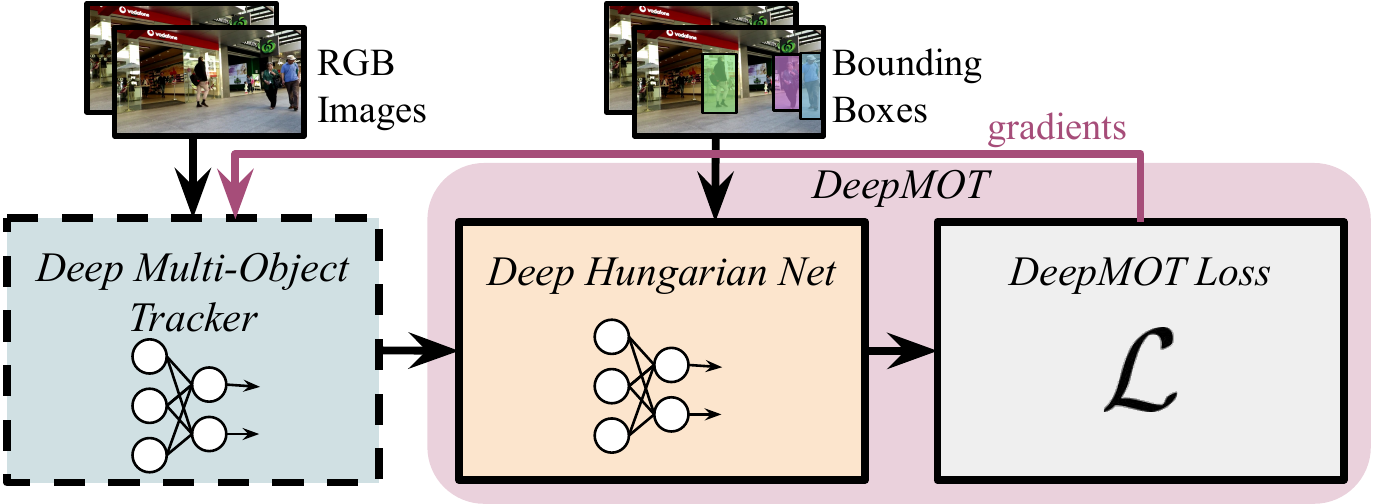}
    \caption{We propose \method, a general framework for training deep multi-object trackers including the DeepMOT loss that directly correlates with established tracking evaluation measures~\cite{Bernardin08JIVP}.
    The key component in our method is the Deep Hungarian Net (DHN) that provides a soft approximation of the optimal prediction-to-ground-truth assignment, and allows to deliver the gradient, back-propagated from the approximated tracking performance measures, needed to update the tracker weights.}
    \label{fig:teaser}
\end{figure}

Recent data-driven trends in MOT leverage the representational power of deep networks for learning identity-preserving embeddings for data association~\cite{LealTaixe16CVPRW, Son17CVPR, Voigtlaender19CVPR}, learning the appearance model of individual targets~\cite{Chu17ICCV, Zhu18ECCV} and learning to regress the pose of the detected targets~\cite{Bergmann19ICCV}.
However, these methods train individual parts of the MOT pipeline using proxy losses (\eg triplet loss~\cite{Schroff15CVPR} for learning identity embeddings), that are only indirectly related to the MOT evaluation measures~\cite{Bernardin08JIVP}. 
The main difficulty in defining loss functions that resemble standard tracking evaluation measures is due to the need of computing the optimal matching between the predicted object tracks and the ground-truth objects. 
This problem is usually solved by using the Hungarian (Munkres) algorithm (HA)~\cite{Kuhn55NRLQ}, which contains non-differentiable operations. 

The significant contribution of this paper is a novel, differentiable framework for the training of multi-object trackers (Fig.~\ref{fig:teaser}): it proposes a differentiable variant of the standard CLEAR-MOT~\cite{Bernardin08JIVP} evaluation measures, which we combine into a novel loss function, suitable for end-to-end training of MOT methods.
In particular, we introduce a differentiable network module -- Deep Hungarian Net (DHN) -- that approximates the HA and provides a soft approximation of the optimal prediction-to-ground-truth assignment. 
The proposed approximation is based on a bi-directional recurrent neural network (Bi-RNN) that computes the (soft) assignment matrix based on the prediction-to-ground-truth distance matrix.
We then express both the MOTA and MOTP~\cite{Bernardin08JIVP} as differentiable functions of the computed (soft) assignment matrix and the distance matrix. Through DHN, the gradients from the approximated tracking performance measures are back-propagated to update the tracker weights. 
In this way, we can train object trackers in a data-driven fashion using losses that directly correlate with standard MOT evaluation measures.
In summary, this paper makes the following contributions:\vspace{-3mm}
\begin{enumerate}[(i)]
    \item We propose novel loss functions that are directly inspired by standard MOT evaluation measures~\cite{Bernardin08JIVP} for end-to-end training of multi-object trackers.\vspace{-3mm}
    \item In order to back-propagate losses through the network, we propose a new network module -- Deep Hungarian Net -- that learns to match predicted tracks to ground-truth objects in a differentiable manner.\vspace{-3mm}
    \item We demonstrate the merit of the proposed loss functions and differentiable matching module by training the recently published Tracktor~\cite{Bergmann19ICCV} using our proposed framework. We demonstrate improvements over the baseline and establish a new state-of-the-art result on MOTChallenge benchmark datasets~\cite{Milan16arxiv, LealTaixe15arxiv}.
\end{enumerate}
\vspace{-5mm}

\section{Related Work}
\vspace{-2mm}
\PAR{Tracking as Discrete Optimization.} With the emergence of reliable object detectors~\cite{DalalTriggs05CVPR, Felzenszwalb08CVPR, Leibe08TPAMI} tracking-by-detection has become the leading tracking paradigm. 
These methods first perform object detection in each image and associate detections over time, which can be performed online via frame-to-frame bi-partite matching between tracks and detections~\cite{Kuhn55NRLQ}. 
As early detectors were noisy and unreliable, several methods search for the optimal association in an offline or batch fashion, often posed as a network flow optimization problem~\cite{Pirsiavash11CVPR, Schulter17CVPR, Zhang08CVPR, ButtCollins13CVPR, Brendel11CVPR}.
    
Alternatively, tracking can be posed as a maximum-a-posteriori (MAP) estimation problem by seeking an optimal set of tracks as a conditional distribution of sequential track states. 
Several methods perform inference using conditional random fields (CRFs)~\cite{Milan14TPAMI, Choi15ICCV, Osep17ICRA}, Markov chain Monte Carlo (MCMC)~\cite{Oh09TAC} or a variational expectation-maximization~\cite{Ba16CVIU,ban2019variational,Ban16ECCV}.
These methods in general, use hand-crafted descriptors for the appearance model, such as color histograms~\cite{Milan14TPAMI, Cho14ICRA}, optical flow based descriptors~\cite{Choi15ICCV} and/or motion models~\cite{Leibe08TPAMI, Osep17ICRA} as association cues. 
Therefore typically only a few parameters are trainable and are commonly learned using grid/random search or tree of parzen window estimators~\cite{Bergstra13ICML, Osep17ICRA}. 
In the case of CRF-based methods, the weights can be trained using structured SVM~\cite{Taskar03NIPS, Wang16IJCV}.

\PAR{Deep Multi-Object Tracking.} Recent data-driven trends in MOT leverage representational power of deep neural networks. 
Xiang~\etal~\cite{Xiang15CVPR} learn track birth/death/association policy by modeling them as Markov Decision Processes (MDP).
As the standard evaluation measures~\cite{Bernardin08JIVP} are not differentiable, they learn the policy by reinforcement learning.

Several existing methods train parts of their tracking methods using losses, not directly related to tracking evaluation measures~\cite{Bernardin08JIVP}.
Kim~\etal~\cite{Kim15ICCV} leverages pre-learned CNN features or a bilinear LSTM~\cite{Kim18ECCV} to learn the long-term appearance model. 
Both are incorporated into (Multiple Hypothesis Tracking) MHT framework~\cite{Reid79TAC}. Other methods~\cite{gan2019self,LealTaixe16CVPRW, Voigtlaender19CVPR, Son17CVPR} learn identity-preserving embeddings for data association using deep neural networks, trained using contrastive~\cite{Hadsell06CVPR}, triplet~\cite{Schroff15CVPR} or quadruplet loss~\cite{Son17CVPR}. At inference time, these are used for computing data association affinities. Approaches by~\cite{Chu17ICCV, Zhu18ECCV} learn the appearance model of individual targets using an ensemble of single-object trackers that share a convolutional backbone. A spatiotemporal mechanism (learned online using a cross-entropy loss) guides the online appearance adaptation and prevents drifts. All these methods are only partially trained, and sometimes in various stages. Moreover, it is unclear how to train these methods to maximize established tracking metrics.

Most similar to our objective, Wang~\etal~\cite{Wang16IJCV} propose a framework for learning parameters of linear cost association functions, suitable for network flow optimization~\cite{Zhang08CVPR} based multi-object trackers. They train parameters using structured SVM. Similar to our method, they devise a loss function, that resembles MOTA:  
the intra-frame loss penalizes false positives (FP) and missed targets while the inter-frame component of the loss penalizes false associations, ID switches, and missed associations. 
However, their loss is not differentiable and is only suitable for training parameters within the proposed min-cost flow framework. Chu~\etal~\cite{Chu_2019_ICCV} propose an end-to-end training framework that jointly learns feature, affinity and multi-dimensional assignment. However, their losses are not directly based on MOTA and MOTP. Schulter~\etal~\cite{Schulter17CVPR} parameterize (arbitrary) cost functions with neural networks and train them end-to-end by optimizing them with respect to the min-flow training objective. Different from~\cite{Schulter17CVPR}, our approach goes beyond learning the association function, and can be used by any learnable tracking method.  

Bergmann~\etal~\cite{Bergmann19ICCV} propose a tracking-by-regression approach to MOT. 
The method is trained for the object detection task using a smooth $L_1$ loss for the bounding box regressor. 
Empirically, their method is able to regress bounding boxes in high-frame rate video sequences with no significant camera motion. 
Apart from the track birth and death management, this approach is fully trainable, and thus it is a perfect method for demonstrating the merit of our training framework. 
Training this approach on a sequence-level data using our proposed loss further improves the performance and establishes a new state of the art on the MOTChallenge benchmark~\cite{LealTaixe15arxiv}.
\vspace{-3mm}
\section{Overview and Notation}
\vspace{-2mm}
\label{sec:overview}

The objective of any MOT method is to predict tracks in a video sequence. Each track $\track^i$ is associated with an identity $i$, and consists of $L_i$ image bounding boxes $\bbox^i_{t_l}\in\R^4$ (2D location and size), $l=1\ldots,L_i$. The task of a multi-object tracker is to accurately estimate the bounding boxes for all identities through time.

At evaluation time, the standard metrics operate frame-by-frame. At frame $t$, the $N_t$ predicted bounding boxes, $\bbox^{i_1}_t,\ldots,\bbox^{i_{N_t}}_t$ must be compared to the $M_t$ ground-truth objects, $\obj^{j_1}_t,\ldots,\obj^{j_{M_t}}_t$. We first need to compute the correspondence between predicted bounding boxes and ground-truth objects.
This is a non-trivial problem as multiple ground-truth boxes may overlap and thus can fit to several track hypotheses. 
In the following we will omit temporal index $t$ to ease the reading. All expressions will be evaluated with respect to time index $t$ unless specified otherwise.

The standard metrics, proposed in~\cite{Bernardin08JIVP}, tackle this association problem using bi-partite matching. First, a prediction-to-ground-truth distance matrix $\distmat \in \R^{N \times M}$,\footnote{The distance matrix $\distmat$ is considered without those objects/tracks that are thresholded-out, \ie, too far from any possible assignment.} $d_{nm} \in [0, 1]$ is computed. 
For vision-based tracking, an intersection-over-union (IoU) based distance is commonly used. Then, the optimal prediction-to-ground-truth assignment binary matrix is obtained by solving the following integer program using the Hungarian algorithm (HA)~\cite{Kuhn55NRLQ}:\vspace{-1mm}
\begin{align*}
  \assigmat^{*} &= \underset{\assigmat \in \{0,1\}^{N\times M}}{\text{argmin}} \sum_{n,m} d_{nm}a_{nm}, \;\;\text{s.t.}\; \sum_m a_{nm} \leq 1, \forall n;\\
 &\sum_n a_{nm} \leq 1, \forall m; \;\; \sum_{m,n}a_{nm} = \min\{N,M\}.
\end{align*}

By solving this integer program we obtain a mutually consistent association between ground-truth objects and track predictions. The constraints ensure that all rows and columns of the assignment should sum to 1, thus avoiding multiple assignments between the two sets. 
After finding the optimal association, $\assigmat^{*}$, we can compute the MOTA and MOTP measures using $\assigmat^{*}$ and $\distmat$:\footnote{Accounting also for the objects/tracks that were left out.}\vspace{-1mm}
\begin{align}
    \text{MOTA} &= 1 - \frac{\sum_t (\fp_t + \fn_t + \ids_t)}{\sum_t M_t}, \\
     \text{MOTP} &= \frac{\sum_t \sum_{n,m} d_{tnm}a^{*}_{tnm}}{\sum_t |\tp_t |},
\end{align}
where $a_{tnm}^*$ is the $(n,m)$-th entry of $\assigmat^*$ at time $t$.
The true positives ($\tp$) correspond to the number of matched predicted tracks and false positives ($\fp$) correspond to the number of non-matched predicted tracks. False negatives ($\fn$) denote the number of ground-truth objects without a match. Finally, to compute ID switches ($\ids$) we need to keep track of past-frame assignments. Whenever the track assigned to a ground truth object changes, we increase the number of $\ids$ and update the assignment structure.

As these evaluation measures are not differentiable, existing strategies only optimize the trackers' hyper-parameters (using,~\eg~random or grid search) that maximize MOTA or MOTP or a combination of both. In their current version, MOTA and MOTP cannot be directly used for tracker optimization with gradient descent techniques.

\section{DeepMOT}
The first step to compute the CLEAR-MOT~\cite{Bernardin08JIVP} tracking evaluation measures is to perform bi-partite matching between the sets of ground-truth objects and of predicted tracks. Once the correspondence between the two sets is established, we can count the number of TP, FN, and IDS needed to express MOTA and MOTP. As the main contribution of this paper, we propose a differentiable loss inspired by these measures, following the same two-step strategy. We first propose to perform a soft matching between the two sets using a differentiable function, parameterized as a deep neural network. Once we establish the matching, we design a loss, approximating the CLEAR-MOT measures, as a combination of differentiable functions of the (soft) assignment matrix and the distance matrix. 
Alternative measures such as IDF1~\cite{Ristani16ECCV} focus on how long the tracker correctly identifies targets instead of how often mismatches occur. However, 
MOTA and IDF1 have a strong correlation. This is reflected in our results -- by optimizing our loss, we also improve the IDF1 measure (see Sec.~\ref{sec:experiments_motc_ablaton}).   
In the following, we discuss both the differentiable matching module (Sec.~\ref{sec:dhn}) and the differentiable version of the CLEAR-MOT measures~\cite{Bernardin08JIVP} (Sec.~\ref{sec:dmot}).

\begin{figure*}[ht]
    \centering
    \includegraphics[width=0.97\linewidth]{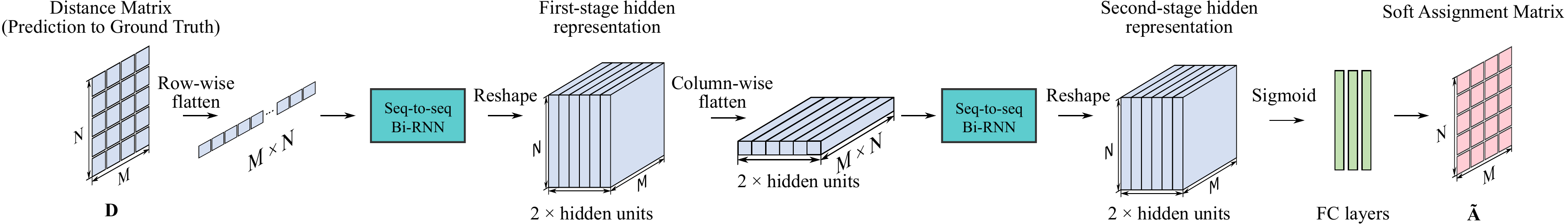}
    \caption{Structure of the proposed Deep Hungarian Net. The row-wise and column-wise flattening are inspired by the original Hungarian algorithm, while the Bi-RNN allows for all decisions to be taken globally, thus is accounting for all input entries.}
    \label{fig:dhn}
\end{figure*}

\subsection{Deep Hungarian Net: DHN}
\label{sec:dhn}

In this section, we introduce DHN, a fundamental block in our DeepMOT framework. DHN produces a proxy $\softassigmat$ that is differentiable w.r.t.\ $\distmat$. Thus DHN provides a bridge to deliver gradient from the loss (to be described later on) to the tracker. We formalize DHN with a non-linear mapping that inputs $\distmat$ and outputs the proxy soft assignment matrix $\softassigmat$. DHN is modeled by a neural network $\softassigmat=g(\distmat,\omega_d)$ with parameters $\omega_d$. Importantly, the DHN mapping must satisfy several properties: (i) the output $\softassigmat$ must be a good approximation to the optimal assignment matrix $\assigmat^*$, (ii) this approximation must be differentiable w.r.t.\ $\distmat$, (iii) both input and output matrix are of equal, but varying size and (iv) $g$ must take global decisions as the HA does.

While (i) will be achieved by setting an appropriate loss function when training the DHN (see Sec.~\ref{sec:experiments_dhn}), (ii) is ensured by designing DHN as a composite of differentiable functions. The requirements (iii) and (iv) push us to design a network that can process variable (but equal) input and output sizes, where every output neuron has a receptive field equals to the entire input. We opt for bi-directional recurrent neural networks (Bi-RNNs). Alternatively, one could consider the use of fully convolutional networks, as these would be able to process variable input/output sizes. However, large assignment problems would lead to partial receptive fields, and therefore, to local assignment decisions.

We outline our proposed architecture in Fig.~\ref{fig:dhn}.
In order to process a 2D distance matrix $\distmat$ using RNNs, we perform row-wise (column-wise) flattening of $\distmat$. This is inspired by the original HA that performs sequentially row-wise and column-wise reductions and zero-entry verifications and fed it to Bi-RNNs (see details below), opening the possibility for $g(\cdot)$ to make global assignment decisions.

Precisely, we perform flattening sequentially, \ie, first row-wise followed by column-wise. The row-wise flattened $\distmat$ is input to a first Bi-RNN that outputs the first-stage hidden representation of size $N\times M\times 2h$, where $h$ is the size of the Bi-RNN hidden layers. Intuitively the first-stage hidden representation encodes the row-wise intermediate assignments. We then flatten the first-stage hidden representation column-wise, to input to a second Bi-RNN that produces the second-stage hidden representation of size $N\times M\times 2h$. The two Bi-RNNs have the same hidden size, but they do not share weights. Intuitively, the second-stage hidden representation encodes the final assignments. To translate these encodings into the final assignments, we feed the second-stage hidden representation through three fully-connected layers (along the $2h$ dimension, \ie, independently for each element of the original $\distmat$). Finally, a sigmoid activation produces the optimal $N\times M$ soft-assignment matrix $\softassigmat$. Note that in contrast to the binary output of the Hungarian algorithm, DHN outputs a (soft) assignment matrix $\softassigmat\in [0, 1]^{N\times M}$.

\PAR{Distance Matrix Computation.}
\label{sec:dist_mat}
The most common metric for measuring the similarity between two bounding boxes is the Intersection-over-Union (IoU). Note that, in principle, the input $\distmat$ can be any (differentiable) distance function. However, if two bounding boxes have no intersection, the distance $1-\text{IoU}$ will always be a constant value of $1$. In that case, the gradient from the loss will be $0$, and no information will be back-propagated.
For this reason, our distance is an average of the Euclidean center-point distance and the Jaccard distance $\mathcal{J}$ (defined as $1-\text{IoU}$):
\vspace{-2mm}
\begin{equation}
    d_{nm} = 
    \frac{
    f(\bbox^n, \obj^m) + \mathcal{J} (\bbox^{n}, \obj^{m})
    }
    {2}.
\end{equation}
$f$ is the Euclidean distance normalized~\wrt the image size:
\begin{equation}
    f(\bbox^n, \obj^m) = 
    \frac{\|c(\bbox^n)-c(\obj^m)\|_2}
    {\sqrt{H^2 + W^2 }},
\end{equation}
where function $c(\cdot)$ computes the center point of the bounding box and $H$ and $W$ are the height and the width of the video frame, respectively.
Both the normalized Euclidean distance and Jaccard distance have values in the range of $[0,1]$, so do all entries $d_{nm}$. %
Our framework admits any distance that is expressed as a composition of differentiable distance functions. In the experimental section, we demonstrate the benefits of adding a term that measures the cosine distance between two learned appearance embeddings. In the following, we explain how we compute a differentiable proxy of MOTA and MOTP as functions of $\distmat$ and $\softassigmat$.

\subsection{Differentiable MOTA and MOTP}
\label{sec:dmot}

In this section, we detail the computation of two components of the proposed DeepMOT loss: differentiable MOTA ($\diffmota$) and MOTP ($\diffmotp$).
As discussed in Sec.~\ref{sec:overview}, to compute the classic MOTA and MOTP evaluation measures, we first find the optimal matching between predicted tracks and ground-truth objects. 
Based on $\assigmat^*$, we count $\fn$, $\fp$ and $\ids$. The latter is computed by comparing assignments between the current frame and previous frames. 
To compute the proposed $\diffmota$ and $\diffmotp$, we need to express all these as  differentiable functions of $\distmat$ and $\softassigmat$ computed using DHN (see Sec.~\ref{sec:dhn}).

\begin{figure}[t]
    \centering
    \includegraphics[width=1.0\linewidth]{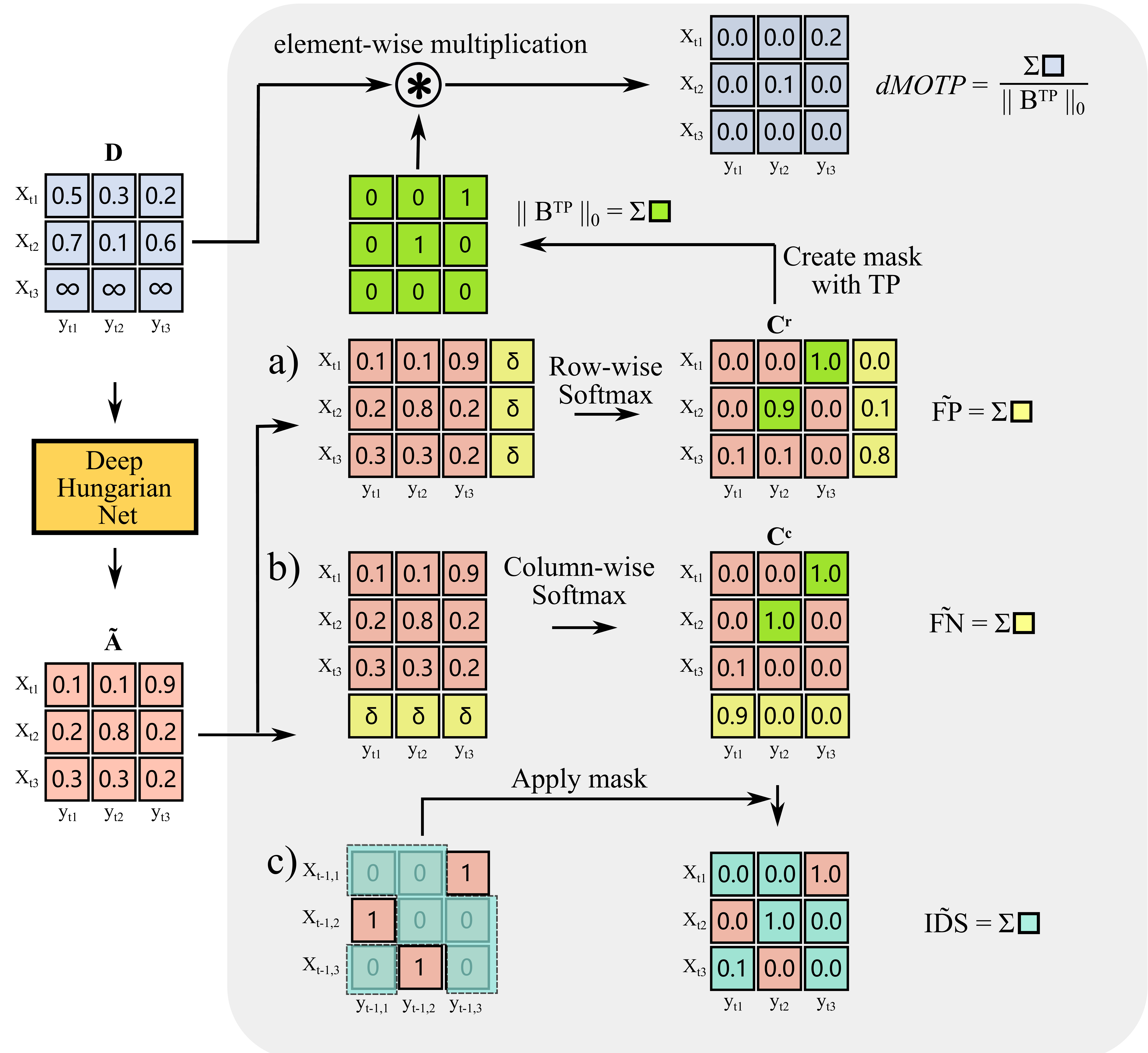}
    \caption{DeepMOT loss: $\diffmotp$ (top) is computed as the average distance of matched tracks and $\diffmota$ (bottom) is composed with $\difffp$, $\diffids$ and $\difffn$.}
    \label{fig:deepmotloss}
\end{figure}

The operations described in the following are illustrated in Fig.~\ref{fig:deepmotloss}. First, we need to count $\fn$ and $\fp$. Therefore, we need to obtain a count of non-matched tracks and non-matched ground-truth objects. 
To this end, we first construct a matrix $\conmat^r$ by appending a column to $\softassigmat$, filled with a threshold value $\delta$ (\eg, $\delta=0.5$), and perform row-wise softmax (Fig.~\ref{fig:deepmotloss}a).
Analogously, we construct $\conmat^c$ by appending a row to $\softassigmat$ and perform column-wise softmax (Fig.~\ref{fig:deepmotloss}b).
Then, we can express a soft approximation of the number of $\fp$ and $\fn$ as:
\begin{equation}
    \difffp = \sum_n \conmat_{  n, M+1}^r, \quad \difffn = \sum_m \conmat_{ N+1, m}^c.
\end{equation}
Intuitively, if all elements in $\softassigmat$ are smaller than the threshold $\delta$, then entries of $\conmat_{n,M+1}^r$ and $\conmat_{N+1, m}^c$ will be close to 1, signaling we have a FP or FN. 
Otherwise, the element with the largest value in each row/column of $\conmat^r$ and $\conmat^c$ (respectively) will be close to 1, signaling that we have a match.
Therefore, the sum of the $N+1$-th row of $\conmat^c$ (Fig.~\ref{fig:deepmotloss}b) and of the $M+1$-th column of $\conmat^r$ (Fig.~\ref{fig:deepmotloss}a) provide an soft estimate of the number of FN and the number of FP, respectively. We will refer to these as $\difffn$ and $\difffp$.

To compute the soft approximations $\diffids$ and $\diffmotp$, we additionally need to construct two binary matrices $\binmat^{\textrm{TP}}$ and $\binmat_{\text{-}1}^{\textrm{TP}}$, whose non-zero entries signal true positives at the current and previous frames respectively. 
Row indices of these matrices correspond to indices assigned to our tracks and column indices correspond to ground truth object identities. We need to pad $\binmat^{\textrm{TP}}_{\text{-}1}$ for element-wise multiplication because the number of tracks and objects varies from frame-to-frame. 
We do this by filling-in rows and columns of $\binmat^{\textrm{TP}}_{\text{-}1}$ to adapt the matrix size for the newly-appeared objects at the current frame by copying their corresponding rows and columns from $\binmat^{\textrm{TP}}$. 
Note that we do not need to modify $\binmat^{\textrm{TP}}$ to compensate for newly appearing objects as these do not cause $\ids$. 
By such construction, the sum of $\conmat_{ 1:N,1:M}^c \odot \overline{\binmat}_{\text{-}1}^{\textrm{TP}}$ (where $\overline{\binmat}$ is the binary complement of $\binmat$) yields the (approximated) number of $\ids$ (Fig.~\ref{fig:deepmotloss}c):
\begin{equation}
\label{eq:diffids}
    \diffids = \| \conmat^c_{1:N,1:M} \odot \overline{\binmat}_{\text{-}1}^{\textrm{TP}} \|_1,
\end{equation}
where $\|\cdot\|_1$ is the $L_1$ norm of a flattened matrix.
With these ingredients, we can evaluate $\diffmota$: %
\begin{equation}
\label{eq:diffmota}
    \diffmota = 1 - \frac{\difffp + \difffn + \gamma\diffids}{M}.
\end{equation}
$\gamma$ controls the penalty we assign to $\diffids$. Similarly, we can express $\diffmotp$ as:
\begin{equation}
\label{eq:diffmotp}
    \diffmotp = 1 - \frac{ \| \distmat \odot \binmat^{\textrm{TP}} \|_1 }{\| \binmat^{\textrm{TP}} \|_0}.
\end{equation}
Intuitively, the $L_1$ norm expresses the distance between the matched tracks and ground-truth objects, and the zero-norm $\|\cdot\|_0$ counts the number of matches.
Since we should train the tracker to maximize MOTA and MOTP, we propose the following DeepMOT loss:
\begin{equation}
\label{eq:deepmotloss}
     {\cal L}_{\textrm{DeepMOT}} = (1 - \diffmota) + \lambda (1 - \diffmotp),
\end{equation}
where $\lambda$ is a loss balancing factor.
By minimizing our proposed loss function ${\cal L}_{\textrm{DeepMOT}}$, we are penalizing $\fp$, $\fn$ and $\ids$ -- all used by the CLEAR-MOT measures~\cite{Bernardin08JIVP}. Same as for the standard CLEAR-MOT measures, $\diffmota$, $\diffmotp$ must be computed at every time frame $t$.
\subsection{How To Train Your Deep Multi-Object Tracker}
\label{sec:howto}
\begin{figure}[t]
    \centering
    \includegraphics[width=1.0\linewidth]{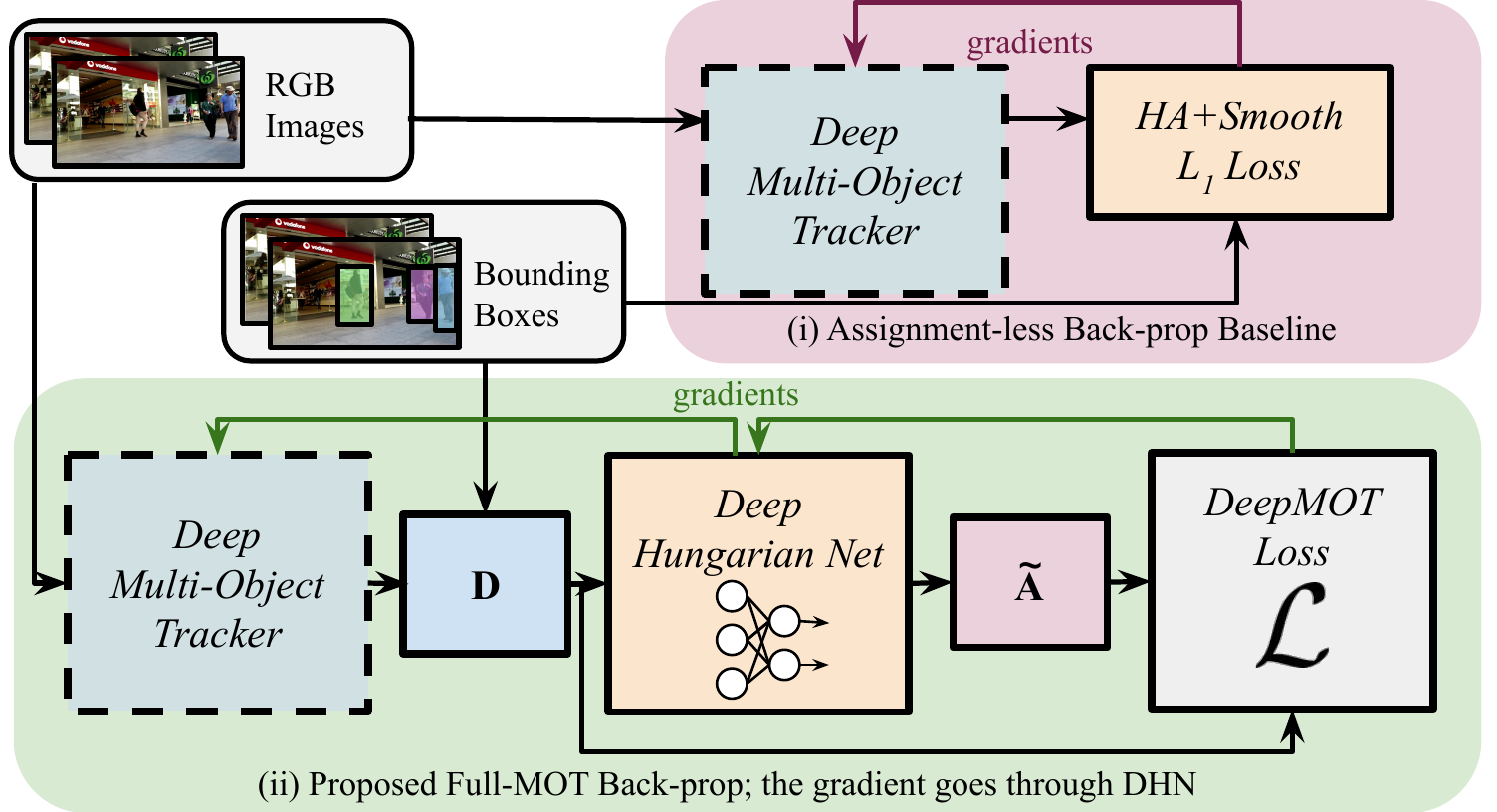}
    \caption{The proposed MOT training strategy (bottom) accounts for the track-to-object assignment problem, solved by the proposed DHN, and approximates the standard MOT losses, as opposed to the classical training strategies (top) using the \textit{non-differentiable} HA.}
    \label{fig:pipeline}
\end{figure}

The overall tracker training procedure is shown in Fig.~\ref{fig:pipeline}.
We randomly sample a pair of consecutive frames from the training video sequences. These two images together with their ground-truth bounding boxes constitute one training instance. 
For each such instance, we first initialize the tracks with ground-truth bounding boxes (at time $t$) and run the forward pass to obtain the track's bounding-box predictions in the following video frame (time $t+1$).
To mimic the effect of imperfect detections, we add random perturbations to the ground-truth bounding boxes (see supplementary material for details).
We then compute $\distmat$ and use our proposed DHN to compute $\softassigmat$ (Sec.~\ref{sec:dhn}). Finally, we compute our proxy loss based on $\distmat$ and $\softassigmat$ (Sec.~\ref{sec:dmot}). This provides us with a gradient that accounts for the assignment, and that is used to update the weights of the tracker.

\section{Experimental Evaluation}
In this section, we first experimentally verify that our proposed DHN is a good approximation to HA~\cite{Kuhn55NRLQ} for bi-partite matching, as required by MOT evaluation measures (Sec.~\ref{sec:experiments_dhn}). 
To show the merit of the proposed framework, we conduct  several experiments on several datasets for evaluating pedestrian tracking performance (Sec.~\ref{sec:experiments_motc}).

\subsection{DHN Implementation Details}
\label{sec:experiments_dhn}

In this section, we provide insights into the performance of our differentiable matching module and outline the training and evaluation details. 

\PAR{DHN Training.} To train the DHN, we create a data set with pairs of matrices ($\distmat$ and $\assigmat^*$), separated into $114,\!483$ matrices for training and $17,\!880$ for matrices testing.
We generate distance matrices $\distmat$ using ground-truth bounding boxes and public detections, provided by the MOT challenge datasets~\cite{Milan16arxiv, LealTaixe15arxiv}. 
We generate the corresponding assignment matrices $\assigmat^*$ (as labels for training) using HA described in \cite{Bernardin08JIVP}.
We pose the DHN training as a 2D binary classification task using the focal loss~\cite{Lin17ICCV}. 
We compensate for the class imbalance (between the number of zeros $n_0$ and ones $n_1$ in $\assigmat^*$) by weighting the dominant zero-class using $w_0 =  n_1 / (n_0+n_1)$.
We weight the one-class by $w_1 = 1- w_0$. We evaluate the performance of DHN by computing the weighted accuracy (WA):
\begin{equation}
    \text{WA} = \frac{w_1 n_1^* + w_0 n_0^*}{w_1n_1 + w_0n_0},
\end{equation}
where $n_1^*$ and $n_0^*$ are the number of true and false positives, respectively. Since the output of the DHN are between $0$ and $1$, we threshold the output at $0.5$. Under these conditions, the network in Fig.~\ref{fig:dhn} scores a WA of $92.88\%$. In the supplementary material, we provide (i) an ablation study on the choice of recurrent unit, (ii) a discussion of alternative architectures, (iii) an analysis of the impact of the distance matrix size on the matching precision and (iv) we experimentally assess how well the DHN preserves the properties of assignment matrices.

\PAR{DHN Usage.} Once the DHN is trained with the strategy described above, its weights are fixed: they are not updated in any way during the training of the deep trackers.

\subsection{Experimental Settings}%
\label{sec:experiments_motc}

We demonstrate the practical interest of the proposed framework by assessing the performance of existing (deep) multi-object trackers when trained using the proposed framework on several datasets for pedestrian tracking. We first ablate the loss terms and the tracking architectures. We also evaluate the impact of the framework with respect to other training alternatives. Finally, we establish a new state-of-the-art score on the MOTChallenge benchmark.

\PAR{Datasets and Evaluation Metrics.} We use the MOT15, MOT16, and MOT17 datasets, which provide crowded pedestrian video sequences captured in the real-world outdoor and indoor scenarios. For the ablation study, we divide the training sequences into training and validation. The details of the split can be found in the supplementary material.
In addition to the standard MOTP and MOTA measures~\cite{Bernardin08JIVP}, we report the performance using the IDF1~\cite{Ristani16ECCV} measure, defined as the ratio of correctly identified detections over the average number of ground-truth objects and object tracks. We also report mostly tracked (MT) and mostly lost (ML) targets, defined as the ratio of ground-truth trajectories that are covered by a track hypothesis more than 80\% and less than 20\% of their life span respectively.

\PAR{Tracktor.} Tracktor~\cite{Bergmann19ICCV} is an adaptation of the Faster RCNN~\cite{Ren15NIPS} object detector to the MOT task. 
It uses a region proposal network (RPN) and the classification/regression heads of the detector to (i) detect objects and (ii) to follow the detected targets in the consecutive frames using a bounding box regression head.
As most parts of Tracktor are trainable, this makes this method a perfect candidate to demonstrate the benefits of our framework.
Note that Tracktor was originally trained only on the MOTChallenge detection dataset and was only applied to video sequences during inference. In the following, we will refer to Tracktor trained in this setting as \textbf{Vanilla Base} Tracktor.
Thanks to DeepMOT, we can train Tracktor directly on video sequences, optimizing for standard MOT measures. We will refer to this variant as \textbf{\method\, Base} Tracktor.

\PAR{Tracktor+ReID.} Vanilla Tracktor has no notion of track identity. 
Therefore~\cite{Bergmann19ICCV} proposed to use an externally trained ReID module during inference to mitigate $\ids$. This external ReID module is a feature extractor with a ResNet-50 backbone, trained using a triplet loss~\cite{Schroff15CVPR} on the MOTChallenge video sequences. We will refer to this variant as \textbf{+ReIDext}.
Note that this does not give Tracktor any notion of identity during training. This means that the DeepMOT loss which penalizes the number of $\ids$ will have no significant effect on the final performance. 
For this reason, we propose to replace \textbf{ReIDext} with a lightweight ReID head that we can train jointly with Tracktor using DeepMOT. 
This in turn allows us to utilize $\diffids$ and to fully optimize performance to all components of CLEAR-MOT measures.  
We refer to this variant as \textbf{+ReIDhead}. It takes the form of a fully-connected layer with $128$ units plugged into  Tracktor. In the supplementary material we provide details on how we embed the ID information into the distance matrix $\distmat$. 

Even if such a network head has been previously used in~\cite{Voigtlaender19CVPR}, it was trained externally using the triplet loss~\cite{Schroff15CVPR}. To the best of our knowledge, we are the first to optimize such an appearance model by directly optimizing the whole network for tracking evaluation measures.

\PAR{MOT-by-SOT.} To demonstrate the generality of our method, we propose two additional simple trainable baselines to perform MOT by leveraging two existing off-the-shelf (trainable) single-object trackers (SOTs): GOTURN~\cite{Held16ECCV} and SiamRPN~\cite{Li18CVPRb}. During inference we initialize and terminate tracks based on object detections.
For each object, the SOTs take a reference image at time $t-1$ of the person and a search region in image $t$ as input. 
Based on this reference box and search region, the SOTs then regress a bounding box for each object independently.

\PAR{Track Management.} In all cases, we use a simple (non-trainable) track management procedure. We (i) use detector responses to initialize object tracks in regions, not covered by existing tracks (can be either public detections or Faster RCNN detection responses in the case of Tracktor); (ii) we regress tracks from frame $t-1$ to frame $t$ using either a SOT or Tracktor and (iii) we terminate tracks that have no overlap with detections (SOT baseline) or invoke the classification head of Tracktor, that signals whether a track is covering an object or not. As an alternative to direct termination, we can set a track as invisible for $K$ frames.

\subsection{Results and Discussion} 
\label{sec:experiments_motc_ablaton}

\begin{table}
\center
\tabcolsep=0.11cm
    \resizebox{\columnwidth}{!}{
    \begin{tabular}{c l c c c c c c c c c}
    \toprule
     & Method & MOTA $\uparrow$ & MOTP $\uparrow$ & IDF1 $\uparrow$ & MT $\uparrow$ & ML $\downarrow$ & FP $\downarrow$ & FN $\downarrow$ & IDS $\downarrow$ \\ [0.5ex] 
     \midrule
     \parbox[t]{3mm}{\multirow{2}{*}{\rotatebox[origin=c]{90}{\footnotesize{Van.}}}} 
        & Base & 59.97 & 89.50 & 70.84 & 35.13 & 27.66 & 276 & 31827 & 326 \\
        & +ReIDext & \textbf{60.20} & \textbf{89.50} &  \textbf{71.15} & \textbf{35.13} & \textbf{27.80}  & \textbf{276}  & \textbf{31827}  & \textbf{152}  \\
        \midrule
        \parbox[t]{3mm}{\multirow{3}{*}{\rotatebox[origin=c]{90}{\footnotesize{\method}}}} 
        & Base & 60.43 & 91.82 & 71.44 & 35.41 & 27.25 & 218 & 31545 & 309 \\
        & +ReIDext & 60.62 & 91.82 & 71.66 & 35.41 & 27.39 & 218 & 31545 & 149 \\ 
        & +ReIDhead & \textbf{60.66} & \textbf{91.82} & \textbf{72.32} & \textbf{35.41} & \textbf{27.25} & \textbf{218} & \textbf{31545} &\textbf{118} \\ 
    \bottomrule
    \end{tabular}
    }

\caption{Impact of the different ReID strategies for the two training strategies on Tracktor's performance.}
\label{tab:mot_ablation_modules}

\end{table}

\PAR{Beyond Bounding Box Regression.} In Tab.~\ref{tab:mot_ablation_modules}, we first establish the Vanilla Base Tracktor performance on our validation set and compare it to the  \method\, Base Tracktor. This experiment (i) validates that our proposed training pipeline based on DHN delivers the gradient to the trackers and improves the overall performance, and (ii) confirms our intuition that training object trackers using a loss that directly correlates with the tracking evaluation measures has a positive impact. Note that the impact on $\ids$ is minimal, which may be on the first sight surprising, as our proposed loss penalizes $\ids$ in addition to FP, FN, and bounding box misalignment.

We study this by first evaluating the impact of applying external ReID module, \ie, \textbf{+ReIDext}. As can be seen in Tab.~\ref{tab:mot_ablation_modules}, \textbf{ReIDext} has a positive impact on the performance, as expected, in terms of MOTA ($+0.23\%$ and $+0.19\%$) and $\ids$ ($-174$ and $-160$) compared to \textbf{Base} for \textbf{Vanilla} and \textbf{DeepMOT} training respectively.
To further demonstrate the interest of a ReID module, we also report the \textbf{+ReIDhead} architecture trained with \method. Importantly, \textbf{+ReIDhead} cannot be trained in the Vanilla setting due to the lack of mechanisms to penalize $\ids$. Remarkably, \textbf{+ReIDhead} trained end-to-end with Tracktor does not only improve over the Base performance (MOTA $+0.23\%$, $\ids$ $\downarrow\!191$), but it also outperforms \textbf{+ReIDext} (MOTA $\uparrow\!0.04$ and $\ids$ $\downarrow\!31$). Very importantly, the lightweight ReID head contains a significantly lower number of parameters ($\approx131$~K) compared to the external ReID module ($\approx25$~M).

Finally, in addition to improve the performance measures for which we optimize Tracktor, DeepMOT consistently improves tracking measures such as IDF1 ($\uparrow\!1.17$ improvement of \textbf{DeepMOT+ReIDhead} over \textbf{Vanilla+ReIDext}). We conclude that (i) training existing trackers using our proposed loss clearly improves the performance and (ii) we can easily extend existing trackers such as Tracktor to go beyond simple bounding box regression and incorporate the appearance module directly into the network. All modules are optimized jointly in a single training.

\begin{table}
\center
\tabcolsep=0.11cm
    \resizebox{\columnwidth}{!}{
    \begin{tabular}{l c c c c c c c c c}
    \toprule
     Training loss & MOTA $\uparrow$ &  MOTP $\uparrow$ & IDF1 $\uparrow$ & MT $\uparrow$ & ML $\downarrow$ & FP $\downarrow$ & FN $\downarrow$ & IDS $\downarrow$ \\ [0.5ex] 
     \midrule
        Vanilla  & 60.20 & 89.50 &  71.15 & 35.13 & 27.80  & 276  & 31827  & 152  \\
        Smooth $L_1$ & 60.38 & 91.81 & 71.27 & 34.99 & 27.25 & 294 & 31649 & 164 \\ 
        \midrule
        $\diffmotp$ & 60.51 & 91.74 & 71.75 & 35.41 & \textbf{26.83} & 291 & 31574 & 142\\
        $\diffmota$ &  60.52 &  88.31 &  71.92 &  35.41 &  27.39 &  254 &  31597 &  142 \\   
        $\diffmota$+$\diffmotp$-$\diffids$ & 60.61  & \textbf{92.03} & 72.10  & 35.41  & 27.25  & 222  & 31579  & 124 \\
        
        $\diffmota$+$\diffmotp$ & \textbf{60.66} & 91.82 & \textbf{72.32} & \textbf{35.41} & 27.25 & \textbf{218} & \textbf{31545} &\textbf{118}  \\

    \bottomrule
    \end{tabular}
    }

\caption{Ablation study on the effect the training loss on Tracktor.}
\label{tab:mot_tracktor_ablation_loss}

\end{table}
    
\PAR{\method\,Loss Ablation.} Next, we perform several experiments in which we study the impact of different components of our proposed loss (Eq.~\ref{eq:deepmotloss}) to the performance of Tracktor (\textbf{\method+ReIDhead}). We outline our results in Tab.~\ref{tab:mot_tracktor_ablation_loss}. In addition to \textbf{Vanilla+ReIDext} (representing the best performance trained in Vanilla settings), we also report results obtained by training the same architecture using only the Smooth $L_1$ loss (see Fig.~\ref{fig:pipeline}). We train the regression head with Smooth $L_1$ loss using a similar training procedure as for \method\,(see Sec.~\ref{sec:howto}), to regress predicted bounding boxes to the ones at current time step of their associated tracks. This approach is limited in the sense that we cannot (directly) penalize FP, FN and IDS. 

The Smooth $L_1$ training, when compared to Vanilla, has a positive impact on almost all performance measures, except for MT, FP, and IDS. However, both Vanilla and Smooth $L_1$ are outperformed almost systematically for all performance measures by the various variants of the DeepMOT loss. Remarkably, when using $\diffmota$ term in our loss, we significantly reduce the number of $\ids$ and FP. Training with $\diffmotp$ has the highest impact on MOTP, as it is the case when training with Smooth $L_1$. When only optimizing for $\diffmota$, we have a higher impact on the MOTA and IDF1 measure. Remarkably, when training with ($\diffmota$+$\diffmotp$), we obtain a consistent improvement on all tracking evaluation measures with respect to Vanilla and Smooth $L_1$. Finally, we asses the impact of $\diffids$, by setting the weight $\gamma$ to 0 (Eq.~\ref{eq:diffmota}) (line $\diffmota$+$\diffmotp$-$\diffids$). In this settings, the trackers exhibits a higher number of $\ids$ compared to using the full loss, confirming that the latter is the best strategy.\\

\PAR{MOT-by-SOT Ablation.} Using DeepMOT, we can turn trainable SOT methods into trainable MOT methods by combining them with the track management mechanism (as explained in Sec.~\ref{sec:experiments_motc}) and optimize their parameters using our loss. In Tab.~\ref{tab:mot_sot_ablation}, we outline the results of the two MOT-by-SOT baselines (GOTURN~\cite{Held16ECCV} and SiamRPN~\cite{Li18CVPRb}). For both, we show the performance when using (i) a pre-trained network, (ii) a network fine-tuned using the Smooth $L_1$ loss, and (iii) the one trained with \method.

\begin{table}
\center
\tabcolsep=0.11cm
    \resizebox{\columnwidth}{!}{
    \begin{tabular}{c l c c c c c c c c c}
    \toprule
     & Training & MOTA $\uparrow$ & MOTP $\uparrow$ & IDF1 $\uparrow$ & MT $\uparrow$ & ML $\downarrow$ & FP $\downarrow$ & FN $\downarrow$ & IDS $\downarrow$ \\ [0.5ex] 
     \midrule
        \parbox[t]{3mm}{\multirow{3}{*}{\rotatebox[origin=c]{90}{\footnotesize{GOTURN}}} }

        & Pre-trained & 45.99 & 85.87 & 49.83  & 22.27 & 36.51 & 2927 & 39271 & 1577 \\        
        & Smooth $L_1$ & 52.28 & 90.56 & 63.53 & \textbf{29.46} & \textbf{34.58} & 2026 & 36180 & 472  \\        
        & \method & \textbf{54.09} & \textbf{90.95} & \textbf{66.09} & 28.63 & 35.13 & \textbf{927} & \textbf{36019} & \textbf{261} \\        
        \midrule
        \parbox[t]{3mm}{\multirow{3}{*}{\rotatebox[origin=c]{90}{\footnotesize{SiamRPN}}}}         
        & Pre-trained & 55.35 & 87.15 &66.95 & 33.61 & \textbf{31.81} & 1907 & 33925 & 356 \\   
        & Smooth $L_1$ & 56.51 & \textbf{90.88} & 68.38 & \textbf{33.75} &32.64 & 925 &  34151 & 167 \\  
        & \method & \textbf{57.16} & 89.32 & \textbf{69.49} & 33.47 & 32.78 & \textbf{889} &\textbf{33667} & \textbf{161} \\ 
        
        \midrule
        \parbox[t]{3mm}{\multirow{3}{*}{\rotatebox[origin=c]{90}{\footnotesize{Tracktor}}}}         
        & Vanilla  & 60.20 & 89.50 &  71.15 & 35.13 & 27.80  & 276  & 31827  & 152  \\
        & Smooth $L_1$ & 60.38 & 91.81 & 71.27 & 34.99 & 27.25 & 294 & 31649 & 164 \\ 
        & \method  & \textbf{60.66} & \textbf{91.82} & \textbf{72.32} & \textbf{35.41} & \textbf{27.25} & \textbf{218} & \textbf{31545} &\textbf{118} \\
    \bottomrule
    \end{tabular}
    }

\caption{DeepMOT vs.~Smooth $L_1$ using MOT-by-SOT baselines and Tracktor.}
\label{tab:mot_sot_ablation}
\end{table}

Based on the results outlined in Tab.~\ref{tab:mot_sot_ablation}, we conclude that training using the Smooth $L_1$ loss improves the MOTA for both SOTs (GOTURN: $+6.29\%$, SiamRPN: $+1.16\%$). Moreover, compared to models trained with Smooth $L_1$ loss,
we further improve MOTA and reduce the number of $\ids$ when we train them using DeepMOT. For GOTURN (SiamRPN), we record a MOTA improvement of $1.81\%$ ($0.65\%$) while reducing the number of IDS by 211 (6). We also outline the improvements comparing \textbf{Vanilla+ReIDext} Tracktor trained with Smooth $L_1$ loss, and \textbf{\method+ReIDhead} Tracktor trained using \method. 
These results further validate the merit and generality of our method for training deep multi-object trackers.

\begin{table}
\center
\tabcolsep=0.11cm

    \resizebox{\columnwidth}{!}{
    \begin{tabular}{c l c c c c c c c c c}
    \toprule
     & Method & MOTA $\uparrow$ & MOTP $\uparrow$ & IDF1 $\uparrow$ & MT $\uparrow$ & ML $\downarrow$ & FP $\downarrow$ & FN $\downarrow$ & IDS $\downarrow$ \\ [0.5ex] 
     \midrule
     
     \parbox[t]{3mm}{\multirow{10}{*}{\rotatebox[origin=c]{90}{MOT17}}}
        & \method-Tracktor & \textbf{53.7} & 77.2 & 53.8  & 19.4  & 36.6  & \textbf{11731}  & 247447 & 1947  \\
        & Tracktor~\cite{Bergmann19ICCV} & 53.5  & 78.0 & 52.3 & 19.5 & 36.6 & 12201 & 248047 & 2072\\\cmidrule{2-10}
        & \method-SiamRPN & 52.1 & \textbf{78.1} & 47.7 & 16.7 & 41.7 & 12132 & 255743 & 2271 \\
        & SiamRPN~\cite{Li18CVPRb} & 47.8 & 76.4 & 41.4 & 17.0 & 41.7 & 38279 & 251989 & 4325 \\\cmidrule{2-10}
        & \method-GOTURN & 48.1 & 77.9 & 40.0 & 13.6 & 43.5 & 22497 & 266515 & 3792 \\
        & GOTURN~\cite{Held16ECCV}  & 38.3 & 75.1 & 25.7 & 9.4 & 47.1 & 55381 & 282670 & 10328 \\\cmidrule{2-10}
        & eHAF~\cite{Sheng18TCVSb} & 51.8 & 77.0 & \textbf{54.7} & \textbf{23.4} & 37.9 & 33212 & \textbf{236772} & 1834 \\
        & FWT~\cite{Henschel17arXiv} & 51.3 & 77.0 & 47.6 & 21.4 & \textbf{35.2} & 24101 & 247921 & 2648 \\
        & jCC~\cite{Keuper18TPAMI} & 51.2 & 75.9 & 54.5   & 20.9 & 37.0 & 25937 & 247822 & \textbf{1802} \\
        & MOTDT17~\cite{Long18ICME} & 50.9 & 76.6 & 52.7 & 17.5 & 35.7 & 24069 & 250768 & 2474 \\
        & MHT\_DAM~\cite{Kim15ICCV} & 50.7 & 77.5 & 47.2 & 20.8 & 36.9 & 22875 & 252889 & 2314 \\
     \midrule
     \midrule
     \parbox[t]{3mm}{\multirow{11}{*}{\rotatebox[origin=c]{90}{MOT16}}}
        & \method-Tracktor & \textbf{54.8} & 77.5 & \textbf{53.4} & \textbf{19.1} & \textbf{37.0} & \textbf{2955} & \textbf{78765} & 645  \\
        & Tracktor~\cite{Bergmann19ICCV} & 54.4 & 78.2 & 52.5 & 19.0 & 36.9 & 3280 & 79149 & 682 \\\cmidrule{2-10} 
        & \method-SiamRPN & 51.8 & 78.1 & 45.5 & 16.1 & 45.1 & 3576 & 83699 & 641 \\
        & SiamRPN~\cite{Li18CVPRb} & 44.0 & 76.6 & 36.6 & 15.5 & 45.7 & 18784 & 82318 & 1047 \\\cmidrule{2-10}
        & \method-GOTURN & 47.2 & 78.0 & 37.2 &13.7 & 46.1 & 7230 & 87781 & 1206 \\
        & GOTURN~\cite{Held16ECCV}  & 37.5 & 75.4 & 25.1 & 8.4 & 46.5 & 17746 & 92867 & 3277 \\\cmidrule{2-10}
        & HCC~\cite{Ma18ACCV} & 49.3 & \textbf{79.0} & 50.7 & 17.8 & 39.9 & 5333 & 86795 & \textbf{391} \\
        & LMP~\cite{Tang17CVPR} & 48.8 &\textbf{79.0} & 51.3 & 18.2 & 40.1 & 6654 & 86245 & 481 \\
        & GCRA~\cite{Ma18ICME} & 48.2 & 77.5 & 48.6 & 12.9 & 41.1 & 5104 & 88586 & 821 \\
        & FWT~\cite{Henschel17arXiv} & 47.8 & 75.5 & 44.3 & \textbf{19.1} & 38.2 & 8886 & 85487 & 852 \\
        & MOTDT~\cite{Long18ICME} & 47.6 &74.8 & 50.9 & 15.2 & 38.3 & 9253 & 85431 & 792 \\
    \bottomrule
    \end{tabular}
    }

\caption{We establish a new state-of-the-art on MOT16 and MOT17 public benchmarks by using the proposed DeepMOT.}
\label{tab:mot}
\end{table}

\PAR{MOTChallenge Benchmark Evaluation} We evaluate the trackers trained using our framework on the MOTChallenge benchmark (test set) using the best-performing configuration, determined previously using the validation set. During training and inference, we use the camera motion compensation module, as proposed by~\cite{Bergmann19ICCV}, for the three trained trackers. We discuss the results obtained on MOT16-17. MOT15 results and parameters are in the supplementary.

We follow the standard evaluation practice and compare our models to methods that are officially published on the MOTChallenge benchmark and peer-reviewed. For MOT16 and MOT17, we average the results obtained using the three sets of provided public detections (DPM~\cite{Felzenszwalb08CVPR}, SDP~\cite{Dollar14TPAMI} and Faster R-CNN~\cite{Ren15NIPS}). %
As in~\cite{Bergmann19ICCV}, we use these public detections for track initialization and termination. Importantly, in the case of Tracktor, we do not use the internal detection mechanism of the network, but only public detections.

As can be seen in Tab.~\ref{tab:mot}, \method-Tracktor establishes a new state-of-the-art on both MOT17 and MOT16.
We improve over Tracktor (on MOT17 and MOT16, respectively) in terms of (i) MOTA ($0.2\%$ and $0.4\%$), (ii) IDF1 ($1.5\%$ and $0.9\%$) and (iii) $\ids$ ($125$ and $37$). On both benchmarks, Vanilla Tracktor is the second best-performing method, and our simple SOT-by-MOT baseline \method-SiamRPN is the third. 
We observe large improvements over our MOT-by-SOT pre-trained models and models trained using \method. For GOTURN, we improve MOTA by $9.8\%$ and $9.7\%$ and we significantly reduce the number of $\ids$ by $6536$ and $2071$, for MOT17 and MOT16 respectively. Similar impact on \method-SiamRPN is observed. 
\section{Conclusion}
In this paper, we propose an end-to-end MOT training framework, based on a differentiable approximation of HA and CLEAR-MOT metrics.
We experimentally demonstrate that our proposed MOT framework improves the performance of existing deep MOT methods. Thanks to our method, we set a new state-of-the-art score on the MOT16 and MOT17 datasets. We believe that our method was the missing block for advancing the progress in the area of end-to-end learning for deep multi-object tracking. We expect that our training module holds the potential to become a building block for training future multi-object trackers.

\subsection*{Acknowledgements}
We gratefully acknowledge the mobility grants from the Department for Science and Technology of the French Embassy in Berlin (SST) and the French Institute for Research in Computer Science and Automation (Inria), especially the Perception team. We are grateful to the Dynamic Vision and Learning Group, Technical University of Munich as the host institute, especially Guillem Bras\'{o} and Tim Meinhardt for the fruitful discussions. Finally, this research was partially funded by the Humboldt Foundation through the Sofja Kovalevskaja Award.

\clearpage
\normalsize
\appendix
\noindent{\LARGE{\textbf{Supplementary Material}}}
\section{Implementation Details}
\subsection{DHN}
\label{subsec:DHNdetails}
For training the DHN, we use the RMSprop optimizer~\cite{Tieleman12slides} with a learning rate of $0.0003$, gradually decreasing by $5\%$ every $20,\!000$ iterations.  
We train DHN for 20 epochs (6 hours on a Titan XP GPU). 
For the focal loss, we weight zero-class by $w_0 =  n_1 / (n_0+n_1)$ and one-class by $w_1 = 1- w_0$. Here $n_0$ is the number of zeros and $n_1$ the number of ones in $\assigmat^*$. 
We also use a modulating factor of $2$ in the focal loss. 
Once the DHN training converges, we freeze the DHN weights and keep them fixed when training trackers with DeepMOT. 

\PAR{Datasets.} To train the DHN, we generate training pairs as follows. We first compute distance matrices $\distmat$ using ground-truth labels (bounding boxes) and object detections provided by the MOTChallenge datasets (MOT 15-17) \cite{LealTaixe15arxiv, Milan16arxiv}.
We augment the data by setting all entries, higher than the randomly (with an uniform distribution ranging from 0 to 1) selected threshold, to a large value to discourage these assignments. This way, we obtain a rich set of various distance matrices. 
We then compute assignments using the (Hungarian algorithm) HA (variant used in~\cite{Bernardin08JIVP}) to get the corresponding (binary) assignment matrices $\assigmat^*$, used as a supervisory signal. 
In this way, we obtain a dataset of matrix pairs ($\distmat$ and $\assigmat^*$), separated into $114,\!483$ training and $17,\!880$ testing instances.

\subsection{Trackers}

\PAR{Datasets.} For training object trackers, we use the MOT17 train set.   
For the ablation studies, we divide the MOT17 into train/val sets. 
We split each sequence into three parts: the first, one containing $50\%$ of frames, the second one $25\%$, and the third $25\%$. 
We use the first $50\%$ for training data and the last $25\%$ for validation to make sure there is no overlap between the two.
In total, we use $2,\!664$ frames for the train set, containing $35,\!836$ ground-truth bounding boxes and 306 identities.
For the validation split, we have $1,\!328$ frames with $200$ identities. The public object detections (obtained by DPM~\cite{Felzenszwalb08CVPR}, SDP~\cite{Yang16CVPR} and Faster RCNN~\cite{Ren15NIPS} detectors) from the MOTChallenge are used only during tracking.

\PAR{Training.} We use the Adam optimizer~\cite{Kingma15ICLR} with a learning rate of $0.0001$. 
We train the SOTs for 15 epochs (72h), and we train Tracktor (regression head and ReID head) for 18 epochs (12h) on a Titan XP GPU.

\PAR{Loss Hyperparameters.} When training trackers using our DeepMOT loss, we set the base value of $\delta=0.5$, and the loss balancing factors of $\lambda=5, \gamma=2$, as determined on the validation set. %

\PAR{Training Details.} To train object trackers, we randomly select one training instance from the sequence that corresponds to a pair of consecutive frames. Then, we initialize object trackers using ground-truth detections and predict track continuations in the next frame. At each time step, we use track predictions and ground-truth bounding boxes to compute $\distmat$, which we pass to our DHN and, finally, compute loss and back-propagate the gradients to the tracker.

\PAR{Data Augmentation.} We initialize trackers using ground-truth bounding boxes. To mimic the effects of imperfect object detectors and prevent over-fitting, we perform the following data augmentations during the training: 
\begin{itemize}
    \item We randomly re-scale the bounding boxes with a scaling factor ranging from $0.8$ to $1.2$. 
    \item We add random vertical and horizontal offset vectors (bounding box width and/or height scaled by a random factor ranging from $0$ to $0.25$).
\end{itemize} 

\PAR{Training with the ReIDhead.} While training Tracktor with our \textbf{ReIDhead}, we make the following changes. 
Instead of selecting a pair of video frames, we randomly select ten consecutive frames. 
This is motivated by the implementation of external ReID mechanism in~\cite{Bergmann19ICCV}, where tracker averages appearance features over ten most recent frames. 
At each training step, we compute representative embedding by averaging embeddings of the past video frames and use it to compute the cosine distance to the ground-truth object embeddings.  

\PAR{Test-time Track Managment.} For the MOT-by-SOT baseline, we use detections from three different detectors (DPM, SDP, and FRCNN) to refine the track predictions.
When the IoU between a track prediction and detection is higher than $0.6$, we output their average.
We also reduce FP in the public detections based on detection scores produced by a Faster RCNN detector. 
For the birth and death processes, we initialize a new track only when detections appear in 3 consecutive frames, and they have a minimal consecutive IoU overlap of $0.3$. 
Tracks that can not be verified by the detector are marked invisible and are terminated after $K=60$ frames. For Tracktor, we use the same track management and suppression strategy as proposed in~\cite{Bergmann19ICCV}.

\section{Additional DHN Ablation}

We perform DHN ablation using our test split of $17,\!880$ DHN training instances, as explained in Sec.~\ref{subsec:DHNdetails}. 
In addition, we evaluate the generalization of DHN by evaluating performing evaluation using distance matrices, generated during the DeepMOT training process. 

\PAR{Accuracy.} We compute the weighted accuracy as (using the same weighting factors $w_1$ and $w_0$ as for the loss):
\begin{equation}
    \text{WA} = \frac{w_1 n_1^* + w_0 n_0^*}{w_1n_1 + w_0n_0}.
\end{equation}\label{eq:wa_supp}
Here, $n_1^*$ and $n_0^*$ are the number of true and false positives, respectively. 

\PAR{Validity.} The output of the matching algorithm should be a permutation matrix; \ie, there should be at most one assignment per row/column. In the case of the HA, this is explicitly enforced via constraints on the solution. 
To study how well the predicted (discretized) assignment matrices preserve this property, we count the number of rows and columns by the following criteria:
\begin{itemize}
    \item \textbf{Several Assignments (SA)} counts the number of rows/columns that have more than one assignment (when performing column-wise maximum and row-wise maximum, respectively).
    \item \textbf{Missing Assignments (MA)} counts the number of rows/columns that are not assigned (when performing column-wise maximum and row-wise maximum, respectively) when ground-truth assignment matrix $\assigmat^{*}$ has an assignment or inversely, no assignment in $\assigmat^{*}$ while $\bar{\assigmat}$ (see below) has an assignment in the corresponding rows/columns. 
\end{itemize}

\PAR{Discretization.} To perform the evaluation, we first need to discretize the soft assignment matrix $\softassigmat$, predicted by our DHN to obtain a discrete assignment matrix $\bar{\assigmat}$. 
There are two possibilities. 
\begin{enumerate}[(i)]
    \item For each row of $\bar{\assigmat}$, we set the entry of $\bar{\assigmat}$ corresponding to the largest value of the row to $1$ (as long as it exceeds $0.5$) and the remaining values are set to $0$. We refer to this variant as \textit{row-wise maximum}.
    \item Analogously, we can perform \textit{column-wise maximum} by processing columns instead of rows.
\end{enumerate}

\begin{figure}[h]
    \centering
    \includegraphics[width=1.0\linewidth]{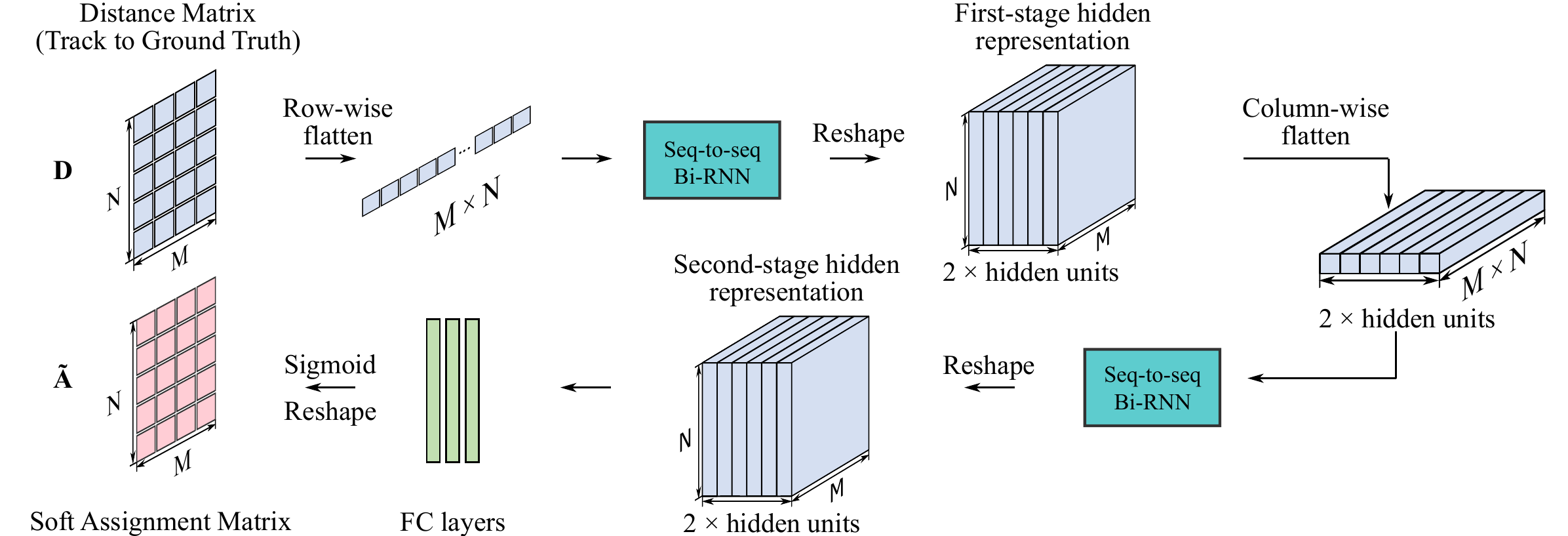}
    \caption{Sequential DHN: Structure of the proposed Deep Hungarian Net. The row-wise and column-wise flattening are inspired by the original Hungarian algorithm, while the Bi-RNN allows for all decisions to be taken globally, thus is accounting for all input entries.}
    \label{fig:dhn_seq}
\end{figure}

\begin{figure}[h]
    \centering
    \includegraphics[width=1.0\linewidth]{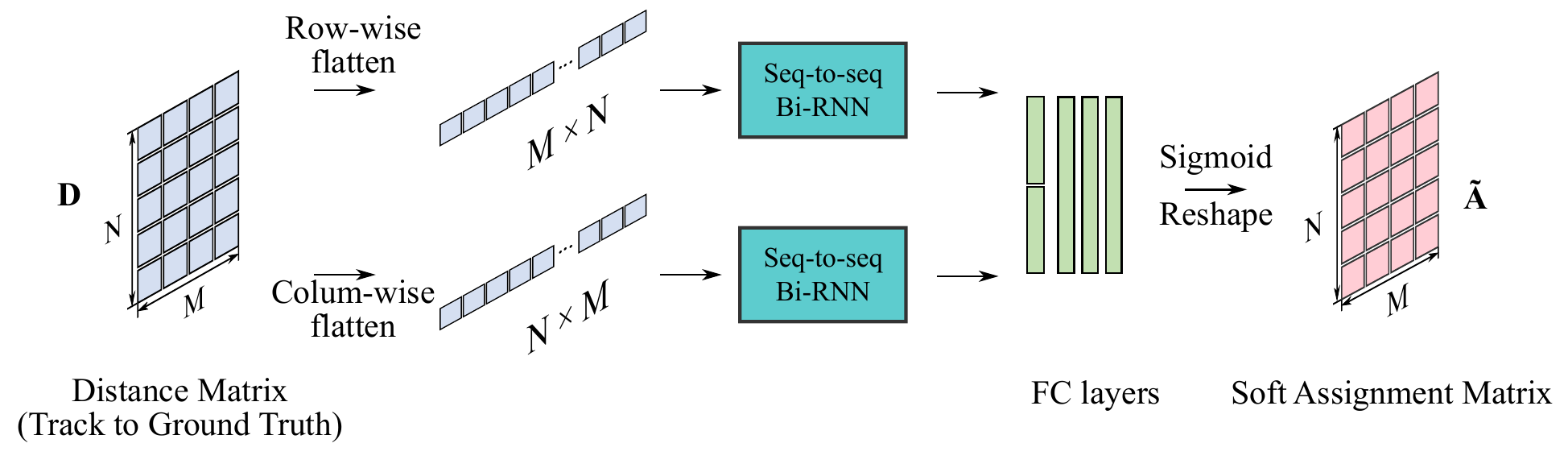}
    \caption{Parallel DHN variant: (i) We perform row-wise and the column-wise flattening of $\distmat$. (ii) We process the flattened vectors using two different Bi-RNNs. (iii) They then are respectively passed to an FC layer for reducing the number of channels and are concatenated along the channel dimension. (iv) After two FC layers we reshape the vector and apply the sigmoid activation.}
    \label{fig:dhn_parallel}
\end{figure}

\begin{figure}[h]
    \centering
    \includegraphics[width=1.0\linewidth]{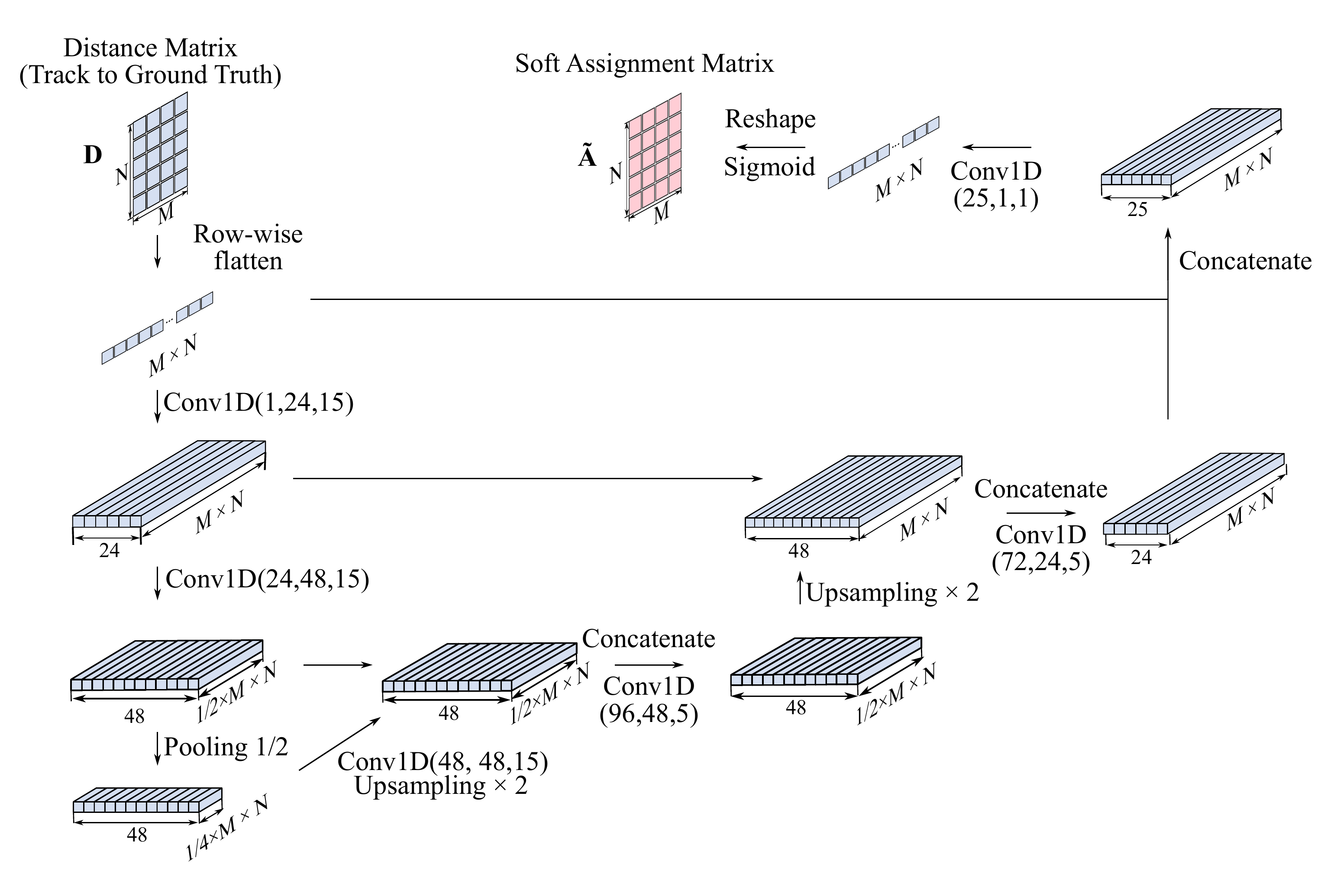}
    \caption{1D convolutional DHN: Our 1D convolutional DHN variant is inspired by the U-Net~\cite{ronneberger2015u}. 
    The encoder consists of two 1D-convolution layers of shapes $[1,24,15]$ and $[24, 48, 15]$ ([\#input channels, \#output channels, kernel size]). 
    The decoder consists of two 1D convolutional layers of shapes $[96,48,5]$ and $[72, 24, 5]$. Finally, we apply an 1D convolution and a sigmoid activation to produce $\softassigmat$.}
    \label{fig:dhn_1dconv}
\end{figure}

\PAR{DHN variants.} We compare three different DHN architectures: 
\begin{enumerate}[(i)]
    \item Sequential DHN (\textbf{seq}, see Fig.~\ref{fig:dhn_seq}),
    \item Parallel DHN (\textbf{paral}, see Fig.~\ref{fig:dhn_parallel}),
    \item 1D Convolutional DHN (\textbf{1d\_conv}, see Fig.~\ref{fig:dhn_1dconv}).
\end{enumerate}
The recurrent unit of the two recurrent architectures, \textbf{seq} and \textbf{paral}, is also ablated between long-short term memory units (\textbf{lstm})~\cite{Hochreiter97NC} and gated recurrent units (\textbf{gru})~\cite{Cho14arxiv}.

\begin{table}
\tabcolsep=0.11cm
\centering
    \begin{center}
     \resizebox{\columnwidth}{!}{
        \begin{tabular}{c|c|ccc}
        \toprule
        Discretization & Network & WA \% ($\uparrow$) & MA\% ($\downarrow$)& SA\% ($\downarrow$) \\
        \midrule
        \multicolumn{1}{c|}{\multirow{5}{*}{\begin{tabular}[c]{@{}c@{}}Row-wise \\ maximum\end{tabular}}}
        &\textbf{seq\_gru (proposed)}& \textbf{92.88}  & \textbf{4.79} & 3.39 \\
        &\textbf{seq\_lstm} & 83.66 & 13.79 &5.98  \\
        &\textbf{paral\_gru}& 89.56  & 8.21 & 4.99 \\
        &\textbf{paral\_lstm}& 88.93 & 8.67& 5.38\\
        &\textbf{1d\_conv}& 56.43 & 35.06& \textbf{2.78}\\ \midrule
        \midrule
        \multicolumn{1}{c|}{\multirow{5}{*}{\begin{tabular}[c]{@{}c@{}}Column-wise \\ maximum\end{tabular}}}
        &\textbf{seq\_gru (proposed)}& \textbf{93.49} & \textbf{6.41} & 26.57 \\
        &\textbf{seq\_lstm} & 87.07 & 13.54 & 47.04\\
        &\textbf{paral\_gru}& 91.01& 7.98& 46.25\\
        &\textbf{paral\_lstm}& 90.50 & 8.60& 47.43 \\
        &\textbf{1d\_conv}& 56.18 & 79.54& \textbf{7.73} \\
        \bottomrule
        \end{tabular}
        }
    \end{center}
\caption{Evaluation results: comparison of different network structures and settings in terms of WA, MA and SA on the DHN test set.}\label{table:results-DHN-testset}
\end{table}

From Tab.~\ref{table:results-DHN-testset}, we see that the proposed sequential DHN (\textbf{seq\_gru}) obtains the highest WA ($92.88\%$ for row-wise maximum and $93.49\%$ for column-wise maximum) compared to others. 
Compared to the 1D convolutional DHN variant (WA of $56.43\%$ and $56.18\%$ for row-wise and column-wise maximum, respectively), Bi-RNN shows the advantage of its global view due to the receptive field, equal to the entire input. 
For the sequential DHN setting, we observe in Tab.~\ref{table:results-DHN-testset} that \textbf{gru} units consistently outperform \textbf{lstm} units with WA $+9.22\%$ (row-wise maximum) and $+6.42\%$ (column-wise maximum). 
Finally, the proposed sequential DHN is more accurate compared to the parallel variant of DHN ($+3.32\%$ for row-wise and $+2.48\%$ for column-wise maximum).
As for the validity, the proposed \textbf{seq\_gru} commits the least missing assignments (MA) ($4.79\%$ and $6.41\%$ for row-wise and column-wise maximum, respectively), and commits only a few SA compared to other variants.

\begin{table}
\centering
\tabcolsep=0.11cm
    \begin{center}
     \resizebox{\columnwidth}{!}{
        \begin{tabular}{c|c|ccc}
        \toprule
        Discretization & Network & WA \% ($\uparrow$) & MA\% ($\downarrow$)& SA\% ($\downarrow$) \\
        \midrule
        \multicolumn{1}{c|}{\multirow{5}{*}{\begin{tabular}[c]{@{}c@{}}Row-wise \\ maximum\end{tabular}}}
        &\textbf{seq\_gru (proposed)}& \textbf{92.71}  & \textbf{13.17} & 9.70 \\
        &\textbf{seq\_lstm} & 91.64 & 14.55 & 10.37  \\
        &\textbf{paral\_gru}& 86.84  & 23.50 & 17.15 \\
        &\textbf{paral\_lstm}& 71.58 & 42.48& 22.62\\
        &\textbf{1d\_conv}& 83.12 & 32.73 & \textbf{5.73} \\ \midrule
        \midrule
        \multicolumn{1}{c|}{\multirow{5}{*}{\begin{tabular}[c]{@{}c@{}}Column-wise \\ maximum\end{tabular}}}
        &\textbf{seq\_gru (proposed)}& \textbf{92.36} & \textbf{12.21} & 3.69 \\
        &\textbf{seq\_lstm} & 91.93 & 13.15 & 4.71\\
        &\textbf{paral\_gru}&87.24 & 20.56 & 16.67\\
        &\textbf{paral\_lstm}& 72.58 & 39.55& 23.16 \\
        &\textbf{1d\_conv}& 82.74 & 32.94 & \textbf{1.11} \\
        \bottomrule
        \end{tabular}
        }
    \end{center}
\caption{Evaluation results. Comparison of different network structures and settings in terms of WA, MA and SA on distance matrices during training.}\label{table:results-DHN-train-matrices}
\end{table}

DHN is a key component of our proposed DeepMOT training framework. 
To evaluate how well DHN performs during training as a proxy to deliver gradients from the DeepMOT loss to the tracker, we conduct the following experiment. 
We evaluate DHN using distance matrices $\distmat$, collected during the DeepMOT training process. 
As can be seen in  Tab.~\ref{table:results-DHN-train-matrices}, the proposed sequential DHN (\textbf{seq\_gru}) outperforms the others variants, with a WA of $92.71\%$ for row-wise and $92.36\%$ for column-wise maximum. 
For validity, it also attains the lowest MA: $13.17\%$ (row) and $12.21\%$ (column). 
The SA is kept at a low level with $9.70\%$ and $3.69\%$ for row-wise and column-wise maximum discretizations, respectively. 
Based on these results, we conclude that (i) our proposed DHN generalizes well to matrices, used to train our trackers, and (ii) it produces outputs that closely resemble valid permutation matrices.

\begin{figure}[h]
    \centering
    \includegraphics[width=1.0\linewidth]{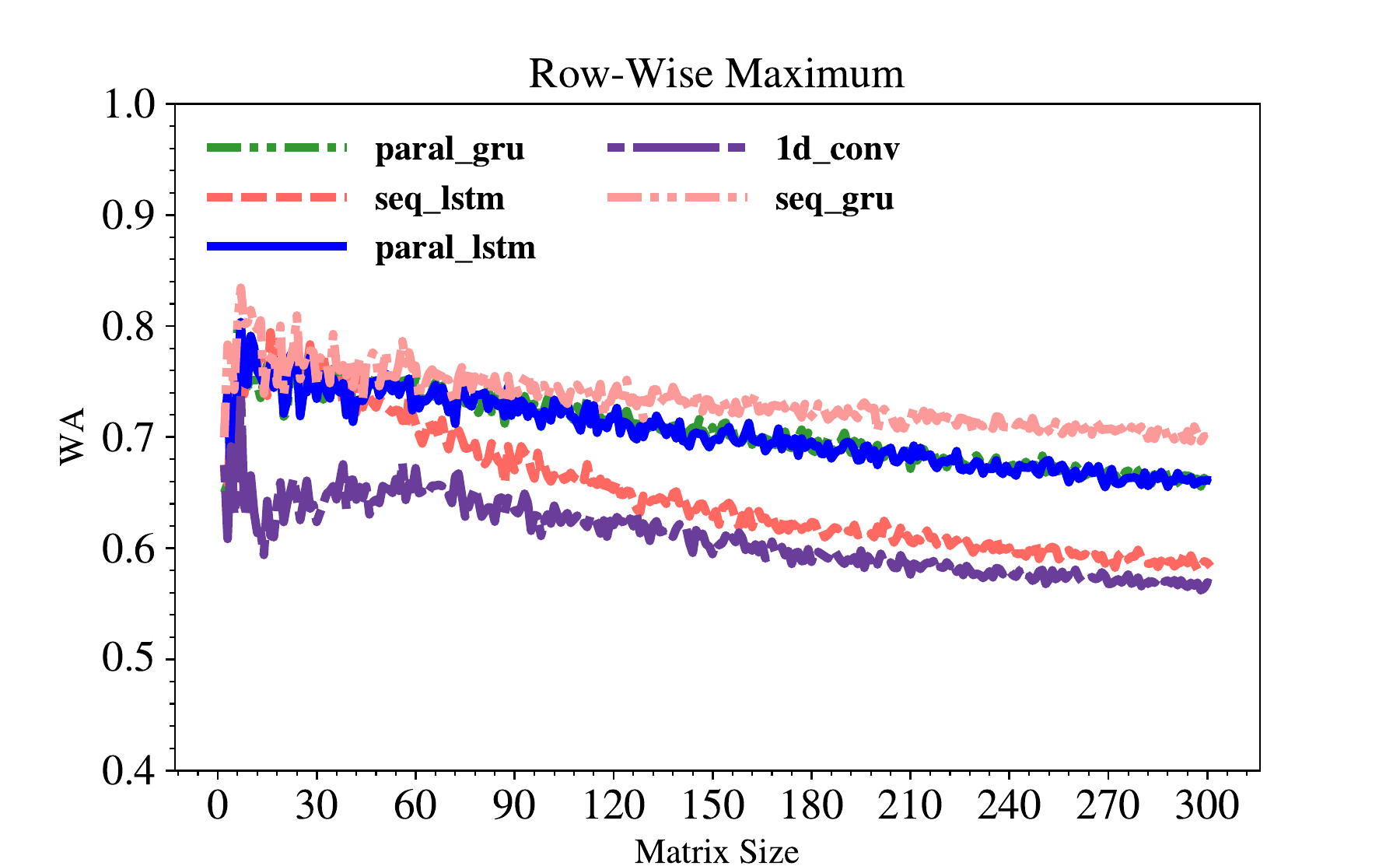}
    \includegraphics[width=1.0\linewidth]{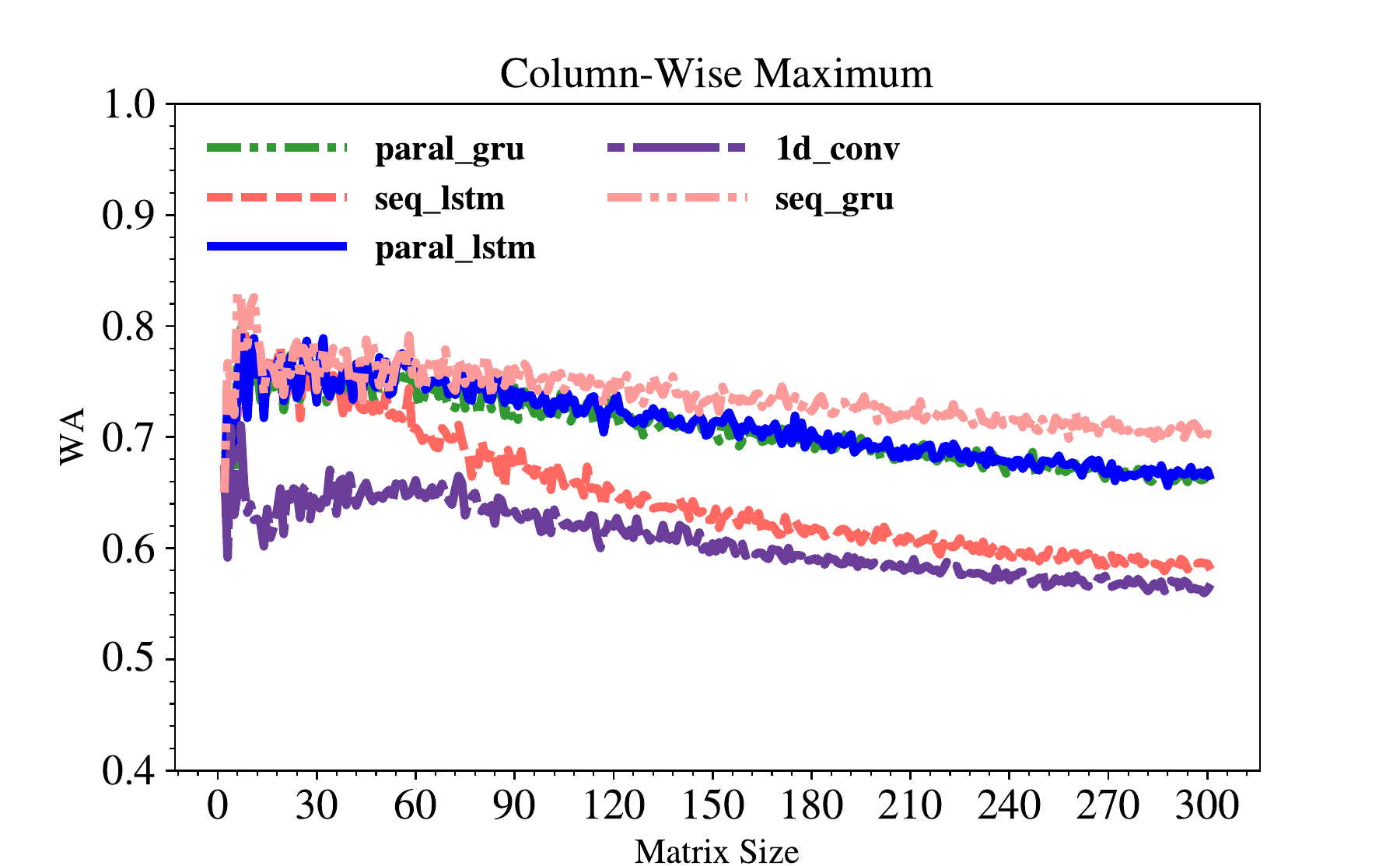}
    \caption{Evaluation of performance of DHN and its variants on $\distmat$ of different sizes.}
    \label{fig:wa_size}
\end{figure}

\PAR{Matrix Size.} To provide further insights into DHN, we study the impact of the distance matrix size on the assignment accuracy.  
We visualize the relation between WA and the input matrix size in Fig.~\ref{fig:wa_size}. 
For validation, we generate square matrices with sizes ranging from $[2, 300]$. 
Precisely, we generate $\distmat$ with a uniform distribution $[0,1)$ and use the Hungarian algorithm implementation from~\cite{Bernardin08JIVP} to generate assignment matrices $\assigmat^*$. 
For each size, we evaluate 10 matrices, which gives us $2,\!990$ matrices in total. %
As can be seen in Fig.~\ref{fig:wa_size}, (i) the proposed \textbf{seq\_gru} consistently outperforms the alternatives. %
(ii) The assignment accuracy of DHN and its variants decreases with the growth of the matrix size. 
Moreover, we observe a performance drop for very small matrices (\ie, $\text{M}=\text{N}\leqslant 6$). 
This may be due to the imbalance with respect to the matrix size during the training. 
\section{Training Gradient Visualization}
\begin{figure}[t!]
    \begin{center}
    \begin{tabular}{cc}
    \includegraphics[width=0.17\textwidth]{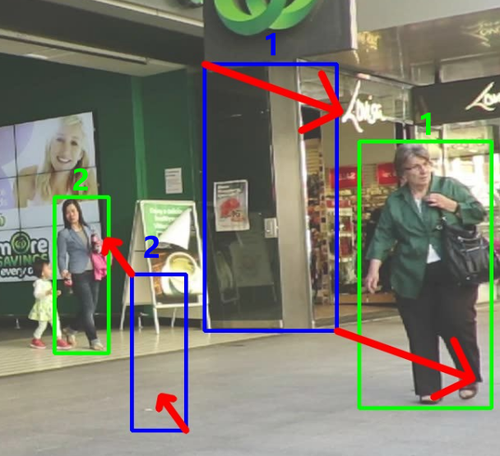} & 
    \includegraphics[width=0.17\textwidth]{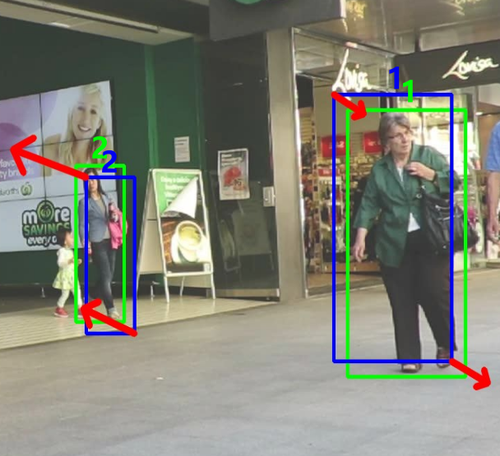} \\ 
    (a) & (b) \\
    \includegraphics[width=0.17\textwidth]{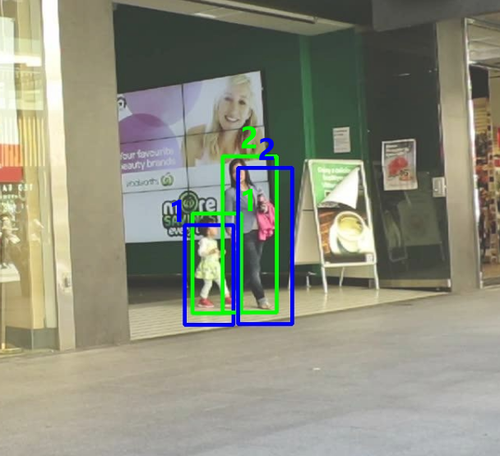} & 
    \includegraphics[width=0.17\textwidth]{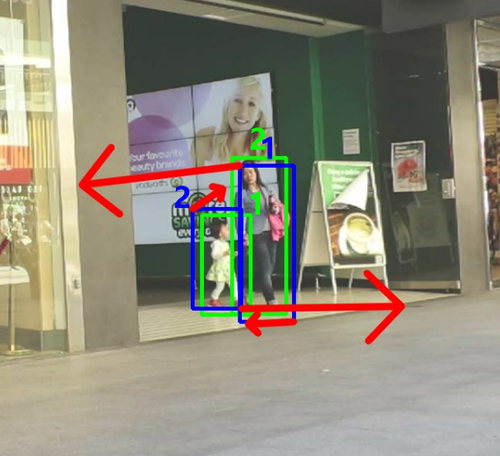} \\
    (c) & (d) \\
    \end{tabular}
    \end{center}
    \caption{Visualization of negative gradients (direction and magnitude) from different terms in the proposed DeepMOT loss: (a) FP and FN (b) MOTP (c-d) IDS (compare (c) $t-1$ with (d) $t$). 
    The predicted bounding-boxes are shown in blue, the ground-truth are shown in green and the gradient direction is visualized using red arrows.}
    \label{fig:gradients}
\end{figure}
The negative gradient should reflect the direction that minimizes the loss. In in Fig.~\ref{fig:gradients} we plot the negative gradient of different terms that constitute our DeepMOT loss \wrt the coordinates of each predicted bounding box to demonstrate visually the effectiveness of our DeepMOT. In this example, we manually generated the cases which contain the FP, FN or IDS. We observe that the negative gradient does encourage the tracks' bounding boxes to be close to those of their associated objects during the training.

\section{MOT15 Results}

\begin{table}
\center
\tabcolsep=0.11cm

    \resizebox{\columnwidth}{!}{
    \begin{tabular}{c l c c c c c c c c c}
    \toprule
     & Method & MOTA $\uparrow$ & MOTP $\uparrow$ & IDF1 $\uparrow$ & MT $\uparrow$ & ML $\downarrow$ & FP $\downarrow$ & FN $\downarrow$ & IDS $\downarrow$ \\ [0.5ex] 
     \midrule
     \parbox[t]{3mm}{\multirow{11}{*}{\rotatebox[origin=c]{90}{2D MOT 2015}}}
        & \method-Tracktor & \textbf{44.1}& \textbf{75.3} & 46.0 & 17.2 & 26.6 & 6085 & 26917 & 1347 \\
        & Tracktor~\cite{Bergmann19ICCV} & \textbf{44.1} & 75.0  & 46.7 & 18.0 & \textbf{26.2} & 6477 & \textbf{26577} & 1318 \\\cmidrule{2-10}
        & \method-SiamRPN & 33.3  & 74.6 & 32.7 & 9.3 & 43.7 & 7825 & 32211 & 919 \\
        & SiamRPN~\cite{Li18CVPRb} & 31.0 & 73.9& 30.7 & 12.6 & 41.7 & 10241 & 31099 & 1062 \\\cmidrule{2-10}
        & \method-GOTURN  & 29.8  & \textbf{75.3} & 27.7 & 4.0 & 66.6 & 3630 & 38964 & 524 \\
        & GOTURN~\cite{Held16ECCV}  & 23.9 & 72.8 & 22.3 & 3.6 & 66.4 & 7021 & 38750 & 965 \\\cmidrule{2-10}
        & AP\_HWDPL\_p~\cite{Chen17ICIP} & 38.5  & 72.6 & \textbf{47.1} & 8.7 & 37.4 & \textbf{4005} & 33203 & 586 \\
        & AMIR15~\cite{Sadeghian17ICCV} & 37.6  & 71.7 & 46.0 & 15.8 & 26.8 & 7933 & 29397 & 1026 \\
        & JointMC~\cite{Keuper18TPAMI} & 35.6  & 71.9& 45.1 & \textbf{23.2} & 39.3 & 10580 & 28508 & 457 \\
        & RAR15pub~\cite{Fang18WACV} & 35.1  & 70.9& 45.4 & 13.0 & 42.3 & 6771 & 32717 & \textbf{381} \\
    \bottomrule
    \end{tabular}
    }

\caption{Results on MOTChallenge MOT15 benchmark.}
\label{tab:mot15}
\end{table}
We summarize the results we obtain on MOT15 dataset in Tab.~\ref{tab:mot15}. Our key observations are: 
\begin{enumerate}[(i)]
    \item For the MOT-by-SOT baseline, we significantly improve over the trainable baselines (SiamRPN and GOTURN). \method-SiamRPN increases MOTA for $+2.3\%$, MOTP for $+0.7\%$ and IDF1 for $+2.0\%$. 
    Remarkably, \method-SiamRPN suppresses $2,\!416$ FP and $143$ IDS. We observe similar performance gains for \method-GOTURN. 
    \item \method-Tracktor obtains results, comparative to the vanilla Tracktor~\cite{Bergmann19ICCV}. Different from MOT16 and MOT17 datasets, we observe no improvements in terms of MOTA, which we believe is due to the fact that labels in MOT15 are very noisy, and vanilla Tracktor already achieves impressive performance. 
    Still, we increase MOTP for $0.3\%$ and reduce FP for $392$. 
\end{enumerate}

{\small
 \bibliographystyle{ieee_fullname.bst}

 }

\end{document}